\documentclass[11pt,a4paper]{article}
\usepackage{times,latexsym}
\usepackage{url}
\usepackage[T1]{fontenc}
\usepackage[utf8]{inputenc}
\usepackage[dvipsnames]{xcolor}
\usepackage{mathtools}
\usepackage{comment}
\usepackage{amsmath}
\usepackage{graphicx}
\usepackage{booktabs}
\usepackage{breqn}
\usepackage{multirow}
\usepackage{enumitem}
\usepackage{breqn}

\usepackage{array}
\usepackage{todonotes}
\usepackage{wrapfig}
\usepackage{arydshln}

\definecolor{mygray}{gray}{0.6}

\newcolumntype{P}[1]{>{\centering\arraybackslash}p{#1}}

\usepackage[acceptedWithA]{tacl2018v2}

\usepackage{xspace,mfirstuc,tabulary}

\newif\iftaclinstructions
\taclinstructionsfalse 
\iftaclinstructions
\renewcommand{\confidential}{}
\renewcommand{\anonsubtext}{(No author info supplied here, for consistency with
TACL-submission anonymization requirements)}
\newcommand{\instr}
\fi

\iftaclpubformat 

\else

\fi

\title{Let's Play {\tt Mono}-{\tt Poly}: BERT Can Reveal Words' Polysemy Level\\ and Partitionability into Senses}

\author{
 Aina Gar\'i Soler \\
 Université Paris-Saclay \\
 CNRS, LISN \\
 91400, Orsay, France \\
  {\tt aina.gari@limsi.fr} \\
  \And
  Marianna Apidianaki \\
  Department of Digital Humanities \\
  University of Helsinki \\
  Helsinki, Finland \\
  {\tt marianna.apidianaki@helsinki.fi} \\
  
}

\date{}

\begin{document}
\maketitle

\begin{abstract}
Pre-trained language models (LMs) encode rich information about linguistic structure but their knowledge about lexical polysemy remains unclear. 
We propose a novel experimental setup for analysing 
this knowledge in LMs specifically trained for different languages  (English, French, Spanish and Greek) and in multilingual BERT.  
We perform our analysis on  datasets carefully designed to reflect different sense distributions, and control for parameters that are highly correlated with polysemy such as frequency and grammatical category. We demonstrate that BERT-derived representations reflect words' polysemy level and their partitionability into senses. Polysemy-related information is more clearly present in English BERT embeddings, but models in other languages also manage to establish relevant distinctions between words at different polysemy levels. 
Our results contribute to a better understanding of the  knowledge encoded in contextualised representations and open up new avenues for multilingual lexical semantics research. 
\end{abstract}

\section{Introduction} \label{sec:intro}

Pre-trained contextual language models have advanced the state of the art in numerous natural language understanding tasks \cite{devlin2019bert,peters2018deep}. Their success has motivated a large number of studies exploring what  these  models  actually learn about language \cite{voita-etal-2019-bottom, 
clark-etal-2019-bert,voita-etal-2019-analyzing,tenney-etal-2019-bert}. 
The bulk of this interpretation work relies on probing tasks which serve to predict linguistic properties from the representations generated by the models \citep{Linzen2018,rogers2020primer}. The focus was initially put on  linguistic aspects pertaining to grammar and syntax \citep{linzen2016assessing,hewitt2019structural,hewitt-liang-2019-designing}. 
The first probing tasks addressing semantic  knowledge explored phenomena in the syntax-semantics interface, such as semantic role labelling and coreference \cite{tenney-etal-2019-bert,kovaleva-etal-2019-revealing}, and the symbolic reasoning potential of LM representations \citep{talmor2019olmpics}. 

  \begin{figure}[t!]
  \centering
\includegraphics[width=.9\columnwidth]{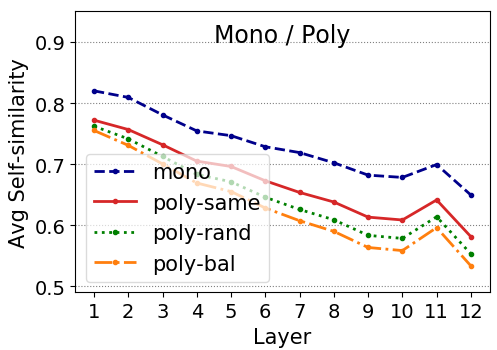}
\caption[width=\textwidth]{BERT distinguishes monosemous ({\tt mono}) from polysemous ({\tt poly}) words 
in all layers. Representations for a {\tt poly} word are obtained from sentences reflecting up to ten different senses ({\tt poly-bal}), the same sense ({\tt poly-same}), or natural occurrence in a corpus ({\tt poly-rand}).} 
 \label{fig:monosemy_ambiguity}
\end{figure}

Lexical meaning was largely overlooked in early interpretation work, but is now attracting increasing attention. 
Pre-trained LMs have been shown to successfully leverage sense annotated data for disambiguation \cite{wiedemann2019does,reif2019visualizing}. The interplay between  
word type and token-level information  
in the hidden representations of  LSTM LMs has also been explored \cite{aina-etal-2019-putting}, as well as  the similarity estimates that can be drawn from  contextualised representations without directly addressing word meaning 
\cite{ethayarajh-2019-contextual}. In recent work, \citet{vulic2020probing} probe BERT representations for lexical semantics, addressing out-of-context word similarity. Whether these models encode knowledge about lexical polysemy and sense distinctions is, however, still an open question. Our work aims to fill this gap. 

We propose methodology for exploring the knowledge about word senses in contextualised representations. Our approach follows a rigorous  experimental protocol proper to  lexical semantic analysis,  which involves the use of datasets carefully designed to reflect different sense distributions. This allows us to investigate the knowledge models acquire during training, and the influence of context variation on  token representations. We account for the strong correlation between word frequency and number of senses \cite{Zipf1945}, and for the relationship between grammatical category and polysemy, by balancing  the frequency and part of speech (PoS) distributions in our datasets and applying a frequency-based model to polysemy prediction.

Importantly, our investigation encompasses  monolingual models in different languages (English, French, Spanish and Greek) and multilingual BERT (mBERT). 
We demonstrate that BERT contextualised representations 
encode an impressive amount of knowledge about polysemy, and are able to 
distinguish monosemous ({\tt mono}) from polysemous ({\tt poly}) words in a variety of settings and configurations (cf. Figure \ref{fig:monosemy_ambiguity}).  
Importantly, we demonstrate that {\bf representations derived from contextual LMs encode knowledge about words' polysemy acquired through pre-training which is combined with information from new contexts of use} (Sections \ref{sec:polysemy_identification}-\ref{sec:classification}). Additionally, we show that {\bf BERT representations can serve to determine how easy it is to partition a word's semantic space into senses} (Section \ref{sec:clusterability}). 

Our methodology can serve for the analysis of words and datasets from different topics, domains and languages. Knowledge  about words' polysemy and sense partitionability has numerous practical implications: it can guide  decisions towards a sense clustering or a  per-instance approach  in applications \cite{reisinger-mooney-2010-multi,neelakantan2014efficient,journals/jair/Camacho-Collados18}; point to words with stable semantics which can be safe cues for disambiguation in running text \cite{leacock-etal-1998-using,agirre-martinez-2004-unsupervised,camachocollados2020}; 
determine the needs in terms of context size 
for disambiguation (e.g. in queries, chatbots); help lexicographers define the number of entries for a word to be present in a resource, and plan the time and effort needed in semantic annotation tasks \cite{mccarthy2016word}.
It could also guide cross-lingual transfer, serving to identify polysemous words for which transfer might be harder. Finally, analysing words' semantic space can be 
highly useful for the study of lexical semantic change \cite{rosenfeld-erk-2018-deep,dubossarsky-etal-2019-time,giulianelli-etal-2020-analysing,schlechtweg2020semeval}. 
We make our code and datasets available 
to enable comparison across studies 
and to promote further research in these directions.\footnote{Our code and data are available at \url{https://github.com/ainagari/monopoly}.}

\section{Related Work}

The knowledge pre-trained contextual LMs encode about lexical semantics 
has only recently started being explored.  
Works by \citet{reif2019visualizing} and \citet{wiedemann2019does} propose experiments using representations built from Wikipedia and the SemCor corpus  \cite{miller-etal-1993-semantic},  and show that BERT can organise word usages in the semantic space in a way that reflects the meaning distinctions present in the data. 
It is also shown that BERT can perform well in the word sense disambiguation (WSD) task  by leveraging the sense-related information available in these resources. 
These works address the disambiguation capabilities of the model but do not show what BERT actually knows about words' polysemy, which is the main axis of our work. In our experiments, sense annotations are {\underline{not}} used to guide the models into establishing sense distinctions, but rather for creating controlled conditions that allow us to analyse BERT's inherent knowledge of lexical polysemy. 

\begin{table*}[t]
    \centering
    \scalebox{0.85}{
    \begin{tabular}{p{1.9cm}p{1.3cm}p{2.3cm}p{11cm}} 
          Setting & Word & Sense &  Sentences \\ 
        \hline
         \multirow{2}{*}{{\tt mono}} & \multirow{2}{*}{hotel.n} & {\sc inn} & The walk ended, inevitably, right in front of his \underline{hotel} building.\\
         & & {\sc inn} & Maybe he's at the \underline{hotel}. \\
         \hline
         \multirow{2}{*}{{\tt poly-same}} & \multirow{2}{*}{room.n} & {\sc chamber} & The \underline{room} vibrated as if a giant hand had rocked it.\\
         & & {\sc chamber} &  (...) Tell her to come to Adam's \underline{room} (...)\\
         \hline
          \multirow{3}{*}{{\tt poly-bal}} & \multirow{3}{*}{room.n} & {\sc chamber} & (...) he left the \underline{room}, walked down the hall  (...)\\ 
        & & {\sc space} & It gives them \underline{room} to play and plenty of fresh air. \\ 
        & & {\sc opportunity} & Even here there is \underline{room} for some variation, for metal surfaces vary (...)\\
     \end{tabular}
     }
    \caption{Example sentences for the monosemous noun {\it hotel} and the polysemous noun {\it room}.} 
    \label{tab:polysemy_sentences}
\end{table*}{}

Probing has also been proposed for lexical semantics analysis, but addressing different questions than the ones posed in our work. \citet{aina-etal-2019-putting} probe the hidden representations of a bidirectional (bi-LSTM) LM for lexical (type-level) and contextual (token-level) information. They specifically train diagnostic classifiers on the tasks of retrieving the input embedding of a word and a representation of its contextual meaning (as reflected in its lexical substitutes).  The results show that the information about the input word that is present in LSTM representations is not lost after contextualisation; however, the quality of the information available for a word is assessed through the model's ability to identify the corresponding embedding, as in \citet{Adi2016FinegrainedAO} and \citet{conneau-etal-2018-cram}. Also, lexical ambiguity is only viewed through the lens of  contextualisation.
In our work, on the contrary, it is given a central role: we explicitly address the knowledge BERT encodes about words' degree of polysemy and partitionability into senses. 
\citet{vulic2020probing} also propose to probe 
contextualised models for lexical semantics, 
but they do so using ``static'' word embeddings obtained through pooling over several contexts, or extracting representations for words in isolation and from BERT's embedding layer, before contextualisation. 
These representations are evaluated on tasks traditionally used for assessing the quality of static embeddings, such as out-of-context similarity and word analogy, which are not tailored for  addressing  lexical polysemy.
Other contemporaneous work explores lexical polysemy in static embeddings \citep{jakubowski-etal-2020-topology}, and the relation of ambiguity and context uncertainty as approximated in the space constructed by mBERT using information-theoretic measures \citep{pimentel-etal-2020-speakers}.
Finally, work by \citet{ethayarajh-2019-contextual} provides useful observations regarding the impact of context on the  representations, without explicitly addressing the semantic knowledge encoded by the models.  
Through an exploration of BERT, ELMo and GPT-2  \citep{radford2019language}, the author highlights the highly distorted similarity of the  obtained contextualised representations which is due to the anisotropy of the vector space built by each model.\footnote{This issue affects all tested models and is particularly present in the last layers of GPT-2, resulting in highly similar representations even for random words.}  
The question of meaning is not addressed in this work, making it hard to draw any conclusions about lexical polysemy. 
 
Our proposed experimental setup is aimed at investigating the polysemy information encoded in the representations built at different layers of deep pre-trained LMs. 
Our approach basically relies on the similarity of contextualised representations, which amounts to  word usage similarity (Usim) estimation, a classical task in lexical semantics \cite{erketal2009,Huang:2012:IWR:2390524.2390645,erk2013measuring}. The Usim task precisely involves predicting the similarity of word instances in context without 
use of sense annotations. BERT has been shown to be particularly good at this task \citep{gari-soler-etal-2019-word,pilehvar-camacho-collados-2019-wic}. Our experiments allow to explore and understand what this ability is due to.

\section{Lexical Polysemy Detection} \label{sec:polysemy_identification}

\subsection{Dataset Creation}
\label{sec:ambigdetect}

We build our English dataset using SemCor 3.0 \cite{miller-etal-1993-semantic}, a corpus manually annotated with WordNet senses \cite{Fellbaum1998}. It is important to note that we  {\underline{do not}}  use the annotations for training or evaluating any of the models. These only serve to control the composition of the sentence pools that are used for generating contextualised representations, and to analyse the results. We form sentence pools for monosemous ({\tt mono}) and polysemous ({\tt poly}) words that occur  at least ten times in SemCor.\footnote{We find the number of senses for a word of a specific part of speech (PoS) in  WordNet 3.0, which we access through the NLTK  interface \cite{bird2009natural}.}
For each {\tt mono} word, we randomly sample ten of its instances in the corpus. 
For each {\tt poly} word, we form three sentence pools of size ten reflecting different sense distributions:  

\begin{itemize}[itemsep=1pt]
     \item \textbf{Balanced} ({\tt poly-bal}).  
     We sample a sentence  \underline{for each sense} of the word in SemCor until a pool of ten  sentences is formed.
    \item \textbf{Random} ({\tt poly-rand}). We randomly sample ten  {\tt poly} word instances from SemCor. We expect this pool to be highly biased towards a specific sense due to  the skewed frequency distribution of word senses  \cite{Kilgarriff2004,mccarthy-etal-2004-finding}.  
    This configuration is closer to the expected natural occurrence of senses in a corpus, it thus serves to estimate the behaviour of the models in a real-world setting. 
    \item \textbf{Same sense} ({\tt poly-same}).  
    We sample ten sentences illustrating {\underline{only one}} 
    sense of the {\tt poly} word. Although the composition of this pool is similar to that of the {\tt mono} pool (i.e. all instances describe the same sense) we call it {\tt poly-same} because it describes one sense of a polysemous word.\footnote{The polysemous words are the same as in {\tt poly-bal} and {\tt poly-rand}.} 
   Specifically, we want to explore whether BERT representations derived from these instances can serve to distinguish {\tt mono} from {\tt poly} words. 
\end{itemize}

\vspace{-1.7mm}

\noindent The controlled composition of the {\tt poly} sentence pools allows us to investigate the behaviour of the models when they are exposed to instances of polysemous words describing the same or different senses. There are 1,765 {\tt poly} words in SemCor with at least 10 sentences available.\footnote{We use sentences of up to 100 words.} We randomly subsample 418 from these in order to balance the {\tt mono} and {\tt poly} classes.
Our English dataset is composed of 836 {\tt mono} and {\tt poly} words, and their instances in 8,195 unique sentences.  
Table \ref{tab:polysemy_sentences} shows a sample of the sentences in different pools. For French, Spanish and Greek, we retrieve sentences from the Eurosense corpus  \cite{bovi2017eurosense} which contains texts from Europarl automatically annotated with BabelNet word senses \cite{NavigliPonzetto:12aij}.\footnote{BabelNet is a multilingual semantic network built from multiple lexicographic and encyclopedic resources, such as WordNet and Wikipedia.} We extract 
sentences from the high precision version\footnote{The high coverage version of Eurosense is larger than the high precision one, but disambiguation is less accurate.} of Eurosense, and create sentence pools in the same way as in
English, balancing the number of monosemous and polysemous words (418). We determine the number of senses for a word as the number of its Babelnet senses that are mapped to a WordNet sense.\footnote{This filtering serves to exclude BabelNet 
senses that correspond to named entities and  
are not useful for our purposes 
(such as movie or album titles), and to run these experiments under similar conditions across languages.}

\subsection{Contextualised Word  Representations}
\label{sec:contextreps}

We experiment with representations generated by three English models: BERT \cite{devlin2019bert}\footnote{We use Huggingface  {\tt transformers} \cite{Wolf2019HuggingFacesTS}.}, ELMo \cite{peters2018deep} and context2vec \cite{melamud2016context2vec}. BERT is a Transformer architecture \cite{vaswani2017attention} that is jointly trained for a masked LM and a next sentence prediction task. 
Masked LM inolves a \textit{Cloze}-style 
task, where the model needs to guess randomly masked words  by jointly conditioning on their left and right context. We use the {\tt bert-base-uncased} and {\tt bert-base-cased} models, pre-trained on the BooksCorpus \citep{Zhuetal2015} and English  Wikipedia. 
ELMo is a bi-LSTM LM trained on Wikipedia and news crawl data from WMT 2008-2012. We use 1024-\textit{d} representations from the 5.5B model.\footnote{\url{https://allennlp.org/elmo}} Context2vec is a neural model based on word2vec's CBOW architecture \cite{mikolov2013efficient}  
which learns embeddings of wide sentential contexts using a bi-LSTM. The model produces representations for words and their context. We use the context representations from a 600-\textit{d} context2vec model pre-trained on the ukWaC corpus \cite{baroni2009wacky}.\footnote{\url{https://github.com/orenmel/context2vec}} 

For French, Spanish and Greek, we use BERT models specifically trained for each language: Flaubert  ({\tt flaubert\_\allowbreak base\_\allowbreak uncased}) \cite{le2020flaubert}, BETO ({\tt bert-\allowbreak base-\allowbreak spanish-\allowbreak wwm-\allowbreak uncased}) \cite{CaneteCFP2020}, and Greek BERT  
({\tt bert-\allowbreak base-\allowbreak greek-\allowbreak uncased-\allowbreak v1}) \cite{koutsikakis2020greek}.  We also use the {\tt bert-base-multilingual-cased} model for each of the four languages. mBERT was trained on Wikipedia data of 104 languages.\footnote{The mBERT model developers recommend using the cased version of the model rather than the uncased one,  especially for languages with non-Latin alphabets, because it fixes normalisation issues. More details about this model can be found here:  \url{https://github.com/google-research/bert/blob/master/multilingual.md}.} All BERT models generate 768-\textit{d} representations.

\subsection{The Self-Similarity Metric}

All models produce representations that describe word meaning in  specific contexts of use. For each instance $i$ of a target word $w$ in a sentence, we extract its representation from: (i) each of the 12 layers of a BERT model;\footnote{We also tried different combinations of the  last four layers, but this did not improve the results. When a word is split into multiple wordpieces (WPs), we obtain its representation by averaging the WPs.} (ii) each of the three ELMo layers; (iii) context2vec. 
We calculate self-similarity ($SelfSim$)  \citep{ethayarajh-2019-contextual} for $w$ in a sentence pool $p$ and a layer $l$, by taking the average of the pairwise cosine similarities of the representations of its instances in $l$: 
\vspace{-6mm}

\begin{equation} \label{eq:selfsim}
SelfSim_{l}(w) = \dfrac{1}{|I|^{2} - |I|} \sum_{i \in I} 
\sum_{\substack{j \in I \\ j \neq i}} cos(x_{wli}, x_{wlj}) 
\end{equation}

\noindent 
In formula \ref{eq:selfsim}, $|I|$ is the number of instances for $w$ (ten in our experiments); $x_{wli}$ and  $x_{wlj}$ are the representations for instances $i$ and $j$ of $w$ in layer $l$. 
The $SelfSim$ score is in the range [-1, 1]. We report the average $SelfSim$ for all $w$'s in a pool $p$. We expect it to be higher 
for monosemous words and words with low polysemy than for highly polysemous words. We also expect the {\tt poly-same} pool to have a higher average $SelfSim$ score than the other {\tt poly} pools which contain instances of different senses. 

Contextualisation has a strong impact on $SelfSim$ since it introduces variation in the token-level representations, making them more dissimilar. The $SelfSim$ value for a word would be 1 with non-contextualised (or static) embeddings, as all its instances would be assigned the same vector. In contextual models, $SelfSim$ is lower in layers where the impact of the context is stronger \citep{ethayarajh-2019-contextual}. It is, however, important to note that contextualisation in BERT models is not monotonic, as shown by previous studies of the models' internal workings  \citep{voita-etal-2019-bottom,ethayarajh-2019-contextual}.
Our experiments presented in the next section provide additional evidence in this respect. 

\begin{figure*}[th!]
    \centering
    \includegraphics[width=\textwidth]{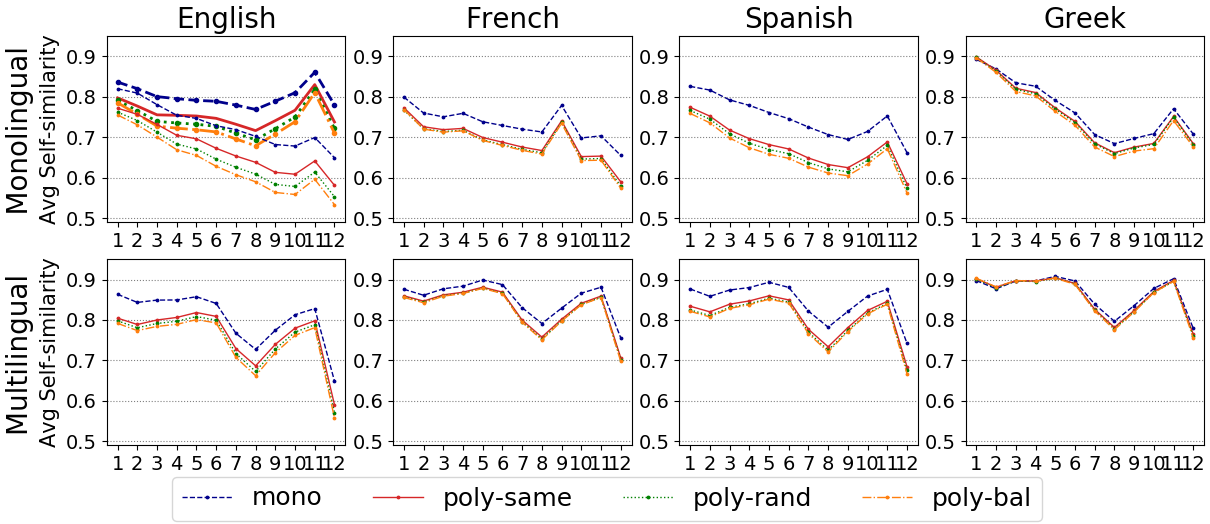}
    \caption{Average $SelfSim$ obtained with monolingual BERT models (top row) and mBERT (bottom row) across all layers 
    (horizontal axis). In the first plot, thick lines  correspond to the {\tt cased} model.}
    \label{fig:multi_monopoly_bert_plots}
\end{figure*}{}

\subsection{Results and Discussion} \label{sec:polysemy_results}

\subsubsection{{\tt Mono-Poly} in English}
\label{sec:sim_english}

Figure \ref{fig:multi_monopoly_bert_plots} shows the average $SelfSim$ value obtained for each sentence pool with representations produced by BERT models. The thin lines in the first plot illustrate the average $SelfSim$ score calculated for {\tt mono} and {\tt poly} words using representations from each layer of the  uncased English BERT model. We observe a clear distinction of words according to their polysemy: $SelfSim$  is higher for {\tt mono} than for {\tt poly} words 
across all layers and sentence pools. BERT establishes a clear distinction even between the {\tt mono} and {\tt poly-same} pools, which contain instances of only one sense.
This distinction is important; it suggests that BERT encodes information about a word's monosemous or polysemous nature regardless of the sentences that are used to derive the contextualised representations. 
Specifically, BERT produces less similar representations for word instances in the {\tt poly-same} pool compared to {\tt mono}, reflecting that {\tt poly} words can have different meanings.

We also observe a clear ordering of the three {\tt poly} sentence pools: average $SelfSim$ is higher in {\tt poly-same}, which only contains instances of one sense, followed by mid-range values in {\tt poly-rand}, and gets its lowest values in the balanced setting ({\tt poly-bal}). This is noteworthy given that  {\tt poly-rand} contains a mix of senses but with a stronger representation of $w$'s most frequent sense than {\tt poly-bal} (71\% vs. 47\%).\footnote{Numbers are macro-averages for words in the pools.} 

\begin{figure}[t!]
    \centering
    \includegraphics[width=\columnwidth]{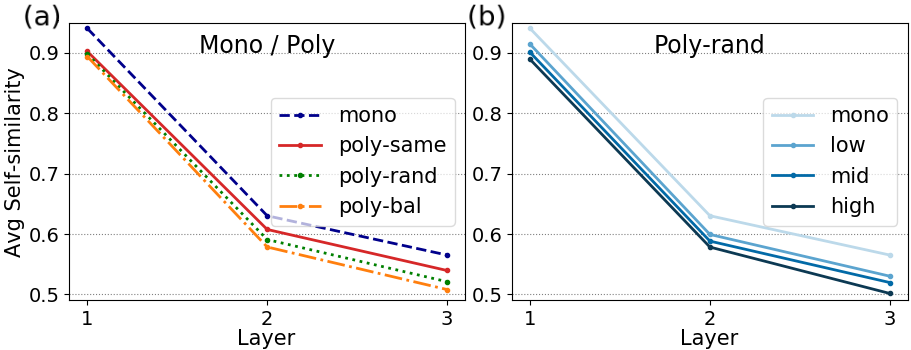}
    \caption{Comparison of average $SelfSim$ obtained for {\tt mono} and {\tt poly} words using ELMo representations (a), and for words in different polysemy bands in the {\tt poly-rand} sentence pool (b).}
    \label{fig:elmo_plots}
\end{figure}

Our results demonstrate  that BERT representations encode two types of lexical semantic knowledge: information about the polysemous nature of words acquired through  pre-training (as reflected in the distinction between {\tt mono} and {\tt poly-same}), and information from the particular instances of a word used to create the  contextualised representations (as shown by the finer-grained distinctions between different {\tt poly} settings).  
BERT's knowledge about polysemy can be due to differences in the types of context where words of different polysemy levels occur. 
We expect  {\tt poly} words to be seen in more varied contexts than {\tt mono} words, reflecting their different senses. BERT encodes this variation with the LM objective through exposure to large amounts of data, and this is reflected in the representations. 
The same ordering pattern is observed with mBERT (lower part of Figure \ref{fig:multi_monopoly_bert_plots}) and with  ELMo (Figure \ref{fig:elmo_plots} (a)). With context2vec, average $SelfSim$ in  {\tt mono} is 0.40, 0.38 in {\tt poly-same}, 0.37 in {\tt poly-rand}, and 0.35 in {\tt poly-bal}. 
This suggests that these models also have some inherent knowledge about lexical polysemy, but differences are less  clearly marked  than in BERT. 

{}

Using the {\tt cased} model leads to an overall increase in $SelfSim$ and to smaller differences between bands, as shown by the thick lines in the first plot of Figure  \ref{fig:multi_monopoly_bert_plots}. Our explanation for the lower distinction ability of the {\tt bert-base-cased} model is that it encodes sparser information about words than the {\tt uncased} model.  It was trained on a more diverse set of strings, so many WPs are present in both their capitalised and non-capitalised form in the vocabulary. In spite of that, it has a smaller vocabulary size (29K WPs) than the uncased model (30.5K). Also, a higher number of WPs correspond to word parts than in the uncased model  (6,478 vs 5,829). 

We test the statistical significance of the {\tt mono}/{\tt poly-rand} distinction using unpaired two-samples t-tests when the normality assumption is met (as determined with Shapiro Wilk's tests). Otherwise, we run a Mann Whitney U test, the non-parametrical alternative of this t-test. In order to lower the probability of type I errors (false positives) that increases when performing multiple tests, we correct p-values using the Benjamini–Hochberg False Discovery Rate (FDR) adjustment \cite{benjamini1995controlling}. Our results show that differences are significant across all embedding types and layers ($\alpha = 0.01$). 

The decreasing trend in $SelfSim$ observed for BERT in Figure  \ref{fig:multi_monopoly_bert_plots}, and the peak in layer 11, confirm the phases of context encoding  and token reconstruction observed by  \citet{voita-etal-2019-bottom}.\footnote{They study the information flow in the Transformer estimating the MI between representations at different layers.} 
In earlier layers, context variation makes representations more dissimilar  and $SelfSim$ decreases. In the last layers, information about the input token is recovered for LM prediction and similarity scores are boosted. Our results show clear distinctions across all BERT and ELMo layers. This suggests that lexical information is spread throughout the layers of the models, and contributes new evidence to the discussion on the localisation of semantic information inside the models
\cite{rogers2020primer,vulic2020probing}.

\begin{figure*}[t!]
  \centering
\includegraphics[width=\textwidth]{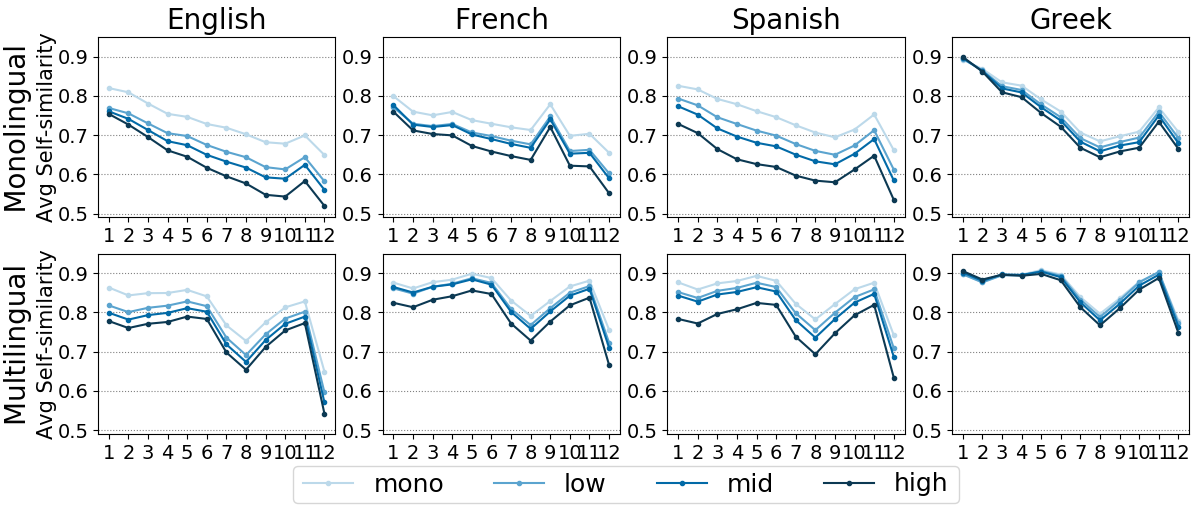}
\caption[width=\textwidth]{Average $SelfSim$ obtained with monolingual BERT models (top row) and mBERT (bottom row) for {\tt mono} and {\tt poly} words in different polysemy bands. Representations are derived from sentences in the {\tt poly-rand} pool.} \label{fig:multi_polysemy_plots}
\end{figure*}

\begin{figure}[t!]
  \centering
\includegraphics[width=\columnwidth]{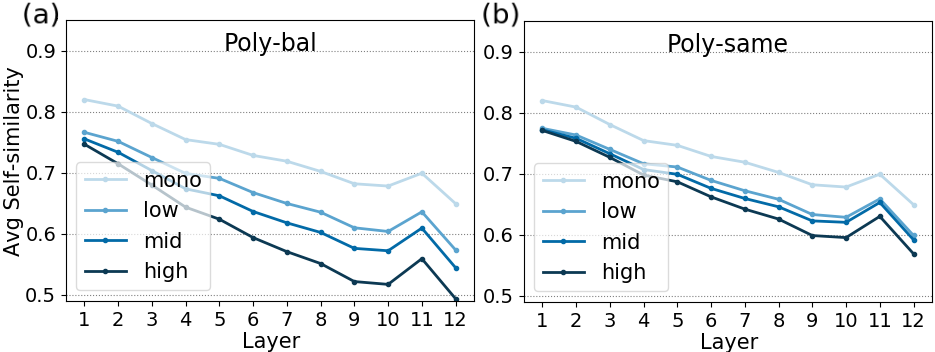}
\caption{Comparison of BERT average $SelfSim$ for {\tt mono} and {\tt poly} words in different polysemy bands in the {\tt poly-bal} and {\tt poly-same} sentence pools.}
 \label{fig:bert_other_polybands_plots}
\end{figure}

\subsubsection{{\tt Mono-Poly} in Other Languages}

The top row of Figure \ref{fig:multi_monopoly_bert_plots} shows the average $SelfSim$ obtained for French, Spanish and Greek words using monolingual models. Flaubert, BETO and Greek BERT representations clearly distinguish {\tt mono} and  {\tt poly} words, but average $SelfSim$ values for different {\tt poly} pools are much closer than in English. BETO  seems to capture these fine-grained distinctions slightly better than the French and Greek models. The second row of the Figure shows results obtained with mBERT representations. We observe the highly similar average $SelfSim$ values assigned to different {\tt poly} pools, which 
show that distinction is harder  than in monolingual models.

Statistical tests show that the difference between $SelfSim$ values in {\tt mono} and {\tt poly-rand} is significant in all layers of BETO, Flaubert, Greek BERT,  and mBERT for Spanish and French.\footnote{In mBERT for Greek, the difference is significant in ten layers.} The magnitude of the difference in Greek BERT is, however, smaller compared to the other models (0.03 vs. 0.09 in BETO at the layers with the biggest difference in average $SelfSim$).

\section{Polysemy Level Prediction}
\label{sec:impactlevelpoly}

\subsection{\textit{SelfSim}-based Ranking} \label{sec:selfsim_based_ranking}

In this set of experiments, we explore the impact of words' degree of polysemy 
on the representations. We control for this factor by grouping words into three polysemy bands as in \citet{mccarthy2016word}, which correspond to a specific number of senses ($k$): {\tt  low}: 2 $\leq$ $k$ $\leq$ 3, {\tt mid}: 4 $\leq$  $k$ $\leq$ 6, {\tt high}: $k$ $>$ 6.  
For English, the three bands are populated with a different number of words: {\tt low}: 551, {\tt mid}: 663, {\tt high}: 551. In the other languages, we form bands containing 300 words each.\footnote{We only used 418 of these {\tt poly} words in Section \ref{sec:polysemy_identification} in order to have balanced {\tt mono} and {\tt poly} pools.}  
In Figure \ref{fig:multi_polysemy_plots}, we compare {\tt mono} words with words in each polysemy band in terms of their average $SelfSim$. Values for {\tt mono} words are taken from Section \ref{sec:polysemy_identification}.
For {\tt poly} words, we use  representations from the {\tt poly-rand} sentence pool which better approximates natural occurrence in a corpus. For comparison, we report in Figure \ref{fig:bert_other_polybands_plots} results obtained in English using sentences from the {\tt poly-same} and {\tt poly-bal} pools.\footnote{We omit the plots for {\tt poly-bal} and {\tt poly-same} for the other models due to space constraints.}  

\begin{figure}[t]
    \centering
    \includegraphics[width=\columnwidth]{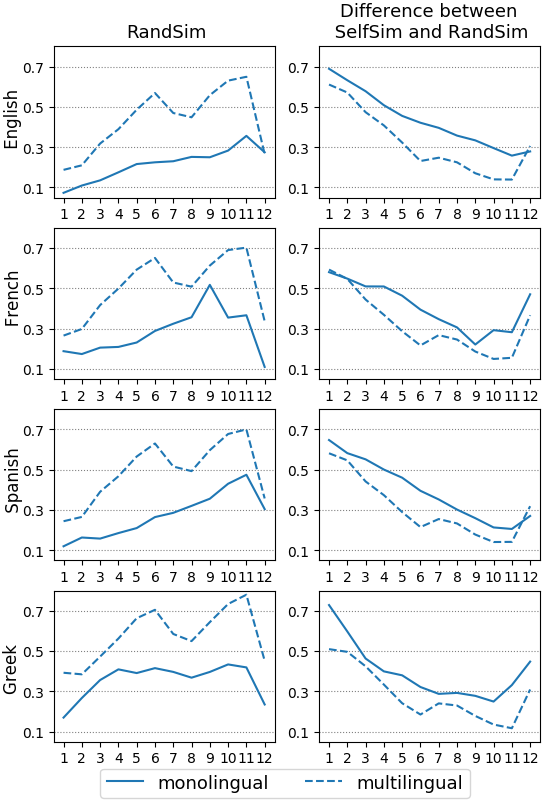}
    \caption{The left plots show the similarity between random words in models for each language. Plots on the right side show the difference between the similarity of random words and $SelfSim$ in {\tt poly-rand}.}
    \label{fig:anisotropy}
\end{figure}

In English, the pattern is clear in all plots: $SelfSim$ is higher for {\tt mono} than for {\tt poly} words in any band, confirming that BERT is able to distinguish {\tt mono} from {\tt poly} words at different polysemy levels. 
The range of $SelfSim$ values for a band is inversely proportional to its $k$: words in {\tt low} get higher values than words in {\tt high}. The results denote that the meaning of highly polysemous words  is more variable (lower $SelfSim$) than the meaning of words with  fewer senses. As expected, scores are higher and inter-band similarities are closer in {\tt poly-same} (cf. Figure \ref{fig:bert_other_polybands_plots} (b)) compared to {\tt poly-rand} and {\tt poly-bal}, where distinctions are clearer. The observed differences confirm that {\bf BERT can predict the polysemy level of words, even from instances describing the same sense}. 

 We observe similar patterns  with ELMo (cf. Figure \ref{fig:elmo_plots} (b)) and context2vec  representations in {\tt poly-rand},\footnote{Average $SelfSim$ values for context2vec in the {\tt poly-rand} setting: {\tt low}: 0.37, {\tt mid}: 0.36, {\tt high}: 0.36.} but smaller absolute inter-band differences.  In {\tt poly-same}, both models fail to correctly order the bands. Overall, our results highlight that  BERT encodes higher quality knowledge about polysemy. We  test the significance of the inter-band  differences detected in {\tt poly-rand} using  the same approach as in Section \ref{sec:sim_english}.  These are  significant in all but a few\footnote{{\tt low}$\rightarrow${\tt mid} in ELMo's third layer, and {\tt mid}$\rightarrow${\tt high} in context2vec and in BERT's first layer.}  layers of the models. 
  
 The bands are also correctly ranked in the other three languages but with smaller  
 inter-band differences than in English, especially in Greek where clear distinctions are only made in a few middle layers. This variation across languages can be explained to some extent by the quality of the automatic EuroSense annotations, which has a direct impact on the quality of the sentence pools. Results of a manual evaluation conducted by \citet{bovi2017eurosense} showed that WSD precision is ten points higher  in English (81.5) and Spanish (82.5) than in French (71.8). The Greek portion, however, has not been evaluated. 
 
 Plots in the second row of Figure \ref{fig:multi_polysemy_plots} show results obtained using mBERT. Similarly to the previous experiment (Section \ref{sec:polysemy_results}),  mBERT overall makes less clear distinctions than the monolingual models. 
 The {\tt low} and {\tt mid} bands often get similar  $SelfSim$ values, which are close to {\tt mono} in French and Greek. Still, inter-band differences are significant in most layers of mBERT and the monolingual French, Spanish and Greek models.\footnote{With the exception of {\tt mono}$\rightarrow${\tt low} in mBERT for 
 Greek, and {\tt low}$\rightarrow${\tt mid} in Flaubert and in mBERT for French.} 

\subsection{Anisotropy Analysis} 

In order to better understand the reasons behind the smaller inter-band differences observed with mBERT, we conduct an additional analysis of the models' anisotropy. We create 2,183 random word pairs from the English {\tt mono}, {\tt low}, {\tt mid} and {\tt high} bands, and 1,318 pairs in each of the other languages.\footnote{1,318 is the total number of words across bands in French, Spanish and Greek.} We calculate the cosine similarity between two random instances of the words in each pair and take the average over all pairs ($RandSim$).  
The plots in the left column  of Figure \ref{fig:anisotropy} show the results. We observe a clear difference in the scores obtained by monolingual models (solid lines) and mBERT (dashed lines). Clearly, mBERT assigns higher similarities to random words, an indication that its semantic space is more anisotropic than the one built by monolingual models. High anisotropy means that representations occupy a narrow cone in the vector space, which results in lower quality similarity estimates and in the model's limited potential to establish clear semantic  distinctions. 

We also compare $RandSim$ to the average $SelfSim$ obtained for {\tt poly} words in the {\tt poly-rand} sentence pool (cf. Section \ref{sec:ambigdetect}). In a quality semantic space, we would expect $SelfSim$ (between same word instances) to be much higher than $RandSim$. The right column of Figure \ref{fig:anisotropy} shows the difference between these two scores. $\mathit{diff}$ in a layer $l$ is calculated as in Equation \ref{eq:selfsim_randsim}: 

\vspace{-6mm}

\begin{equation} \label{eq:selfsim_randsim}
\resizebox{.97\hsize}{!}{$\mathit{diff_l} = \mathrm{Avg}\hspace{0.5mm}SelfSim_l(\mathtt{poly\mathrm{-}rand}) - RandSim_l$}
\end{equation}

\noindent We observe that the difference is smaller in the space built by mBERT, which is more anisotropic than the space built by monolingual models. 
This is particularly obvious in the upper layers of the model. This result confirms the lower quality of mBERT's semantic space compared to monolingual models. 

Finally, we believe that another factor behind the worse mBERT performance is that the multilingual WP vocabulary is mostly English-driven, resulting in arbitrary partitionings of words in the other languages. This word splitting procedure must have an impact on the quality of the lexical information in mBERT representations.

\begin{figure}[t]
    \centering
    \includegraphics[width=\columnwidth]{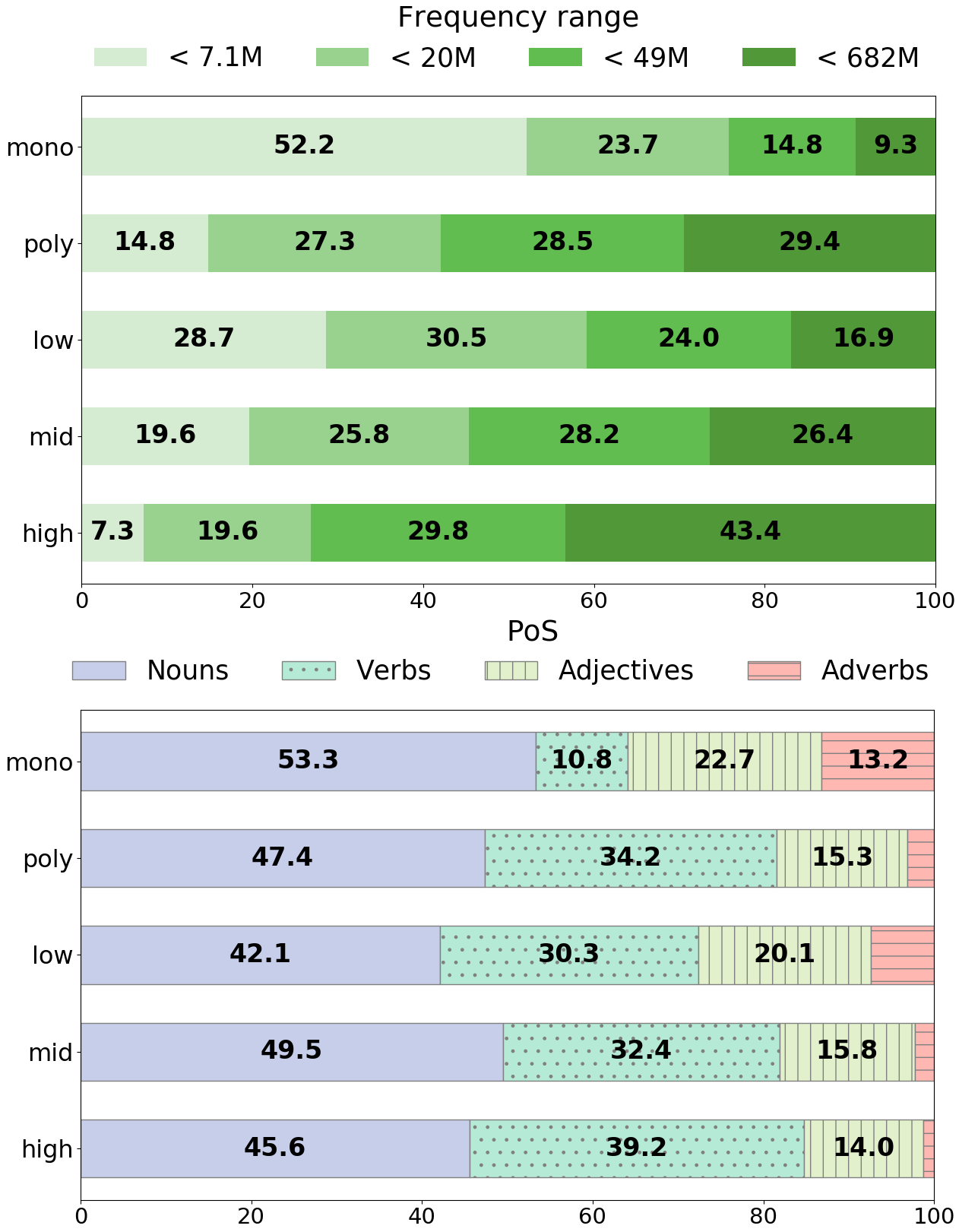}
    \caption{Composition of the English word bands in terms of frequency (a) and grammatical category (b).}
    \label{fig:freqpos_by_band}
\end{figure}{}

 \begin{figure}[t]
    \centering
    \includegraphics[width=\columnwidth]{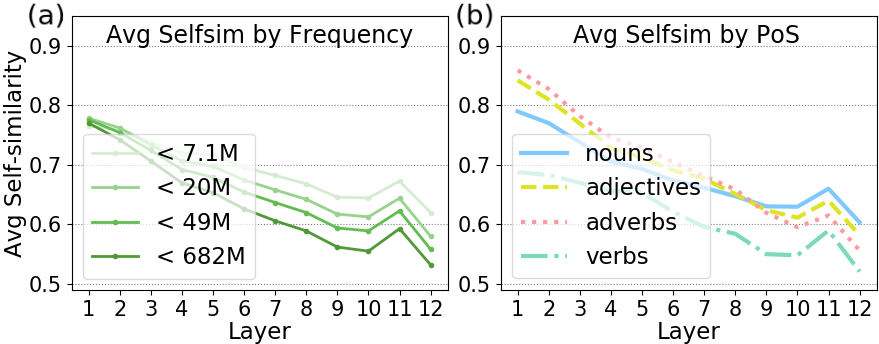}
    \caption{Average $SelfSim$ for English words of different frequencies and part of speech categories with BERT representations.}
    \label{fig:bert_pos_freq_plots}
\end{figure}{}

\section{Analysis by Frequency and PoS} \label{sec:posfreq_analysis}

Given the strong correlation between word frequency and number of senses \cite{Zipf1945}, we explore the impact of frequency on BERT representations. Our goal is to determine the extent to which it influences the good {\tt mono/poly} detection results obtained in Sections \ref{sec:polysemy_results} and \ref{sec:selfsim_based_ranking}.

\subsection{Dataset Composition} 

We perform this analysis in English using frequency information from Google Ngrams  \cite{brants2006web}. For French, Spanish and Greek, we use frequency counts  gathered from the OSCAR corpus \cite{suarez2019asynchronous}. We split the words into four ranges ($F$)  corresponding to the quartiles of frequencies in each dataset. Each range $f$ in $F$ contains the same number of words. We provide detailed information about the composition of the English dataset in Figure \ref{fig:freqpos_by_band}.\footnote{The composition of each band is the same as in Sections \ref{sec:polysemy_identification} and \ref{sec:impactlevelpoly}.} 
Figure \ref{fig:freqpos_by_band}(a) shows that {\tt mono} words are much less frequent than {\tt poly} words. Figure \ref{fig:freqpos_by_band}(b) shows the distribution of different PoS categories in each band. Nouns are the prevalent category in all bands and verbs are less present among {\tt mono} words (10.8\%), as expected.  Finally, adverbs are hardly represented in the {\tt high} polysemy band (1.2\% of all words).

 \begin{figure*}
     \centering 
     \includegraphics[width=\textwidth]{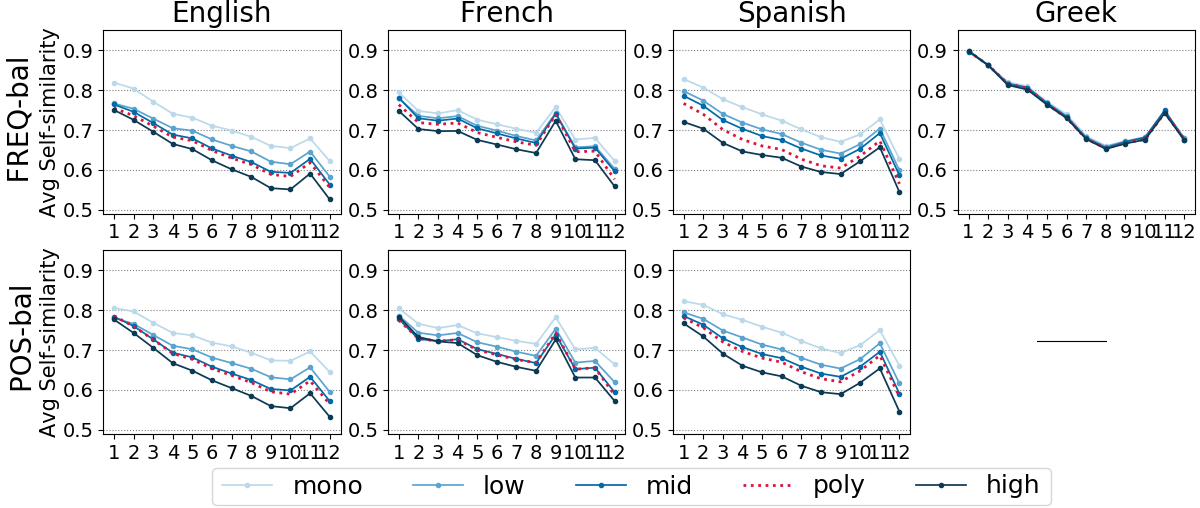}
     \caption{Average $SelfSim$ inside the {\tt poly} bands balanced for frequency ({\sc freq}-bal) and part of speech ({\sc pos}-bal). $SelfSim$ is calculated using representations generated by monolingual BERT models from sentences in each language-specific pool.  We do not balance the Greek dataset for PoS because it only contains nouns.}
     \label{fig:posfreqbalanced_all_langs_mono}
 \end{figure*}

\subsection{\textit{Self-Sim} by Frequency Range  and PoS} 

We examine the average BERT $SelfSim$ per frequency range in {\tt poly-rand}. Due to space constraints, we only report detailed results for the English BERT model in Figure \ref{fig:bert_pos_freq_plots} (plot (a)). 
The clear ordering by range  suggests that BERT can successfully  distinguish words by their frequency, especially in the last layers. 
Plot (b) in Figure  \ref{fig:bert_pos_freq_plots} shows the average $SelfSim$ for words of each PoS category. Verbs  have the lowest $SelfSim$ which is not surprising given that they are highly  polysemous (as shown in Figure  \ref{fig:freqpos_by_band}(b)).
We observe the same trend for monolingual models in the other three languages.

\subsection{Controlling for Frequency and PoS} 

We conduct an additional experiment where we control for the composition of the {\tt poly} bands in terms of grammatical category and word frequency. We call these two settings {\sc pos}-bal and {\sc freq}-bal.  
We define $n_{pos}$, the smallest number of words of a specific PoS that can be found in a band. We form the {\sc pos}-bal bands by subsampling from each band the same number of words ($n_{pos}$) of that PoS. For example,  all {\sc pos}-bal bands have $n_n$ nouns and $n_v$ verbs. We follow a similar procedure to balance the bands by frequency in the {\sc freq}-bal setting. In this case, $n_f$ is the minimum number of words of a specific frequency range $f$ that can be found in a band. We form the {\sc freq}-bal dataset by subsampling from each band the same number of words ($n_f$) of a given range $f$ in $F$. 
 
Table \ref{tab:posfreq_dist_afterbalancing} shows the distribution of words per PoS and frequency range in the {\sc pos}-bal and {\sc freq}-bal settings for each language. The table reads as follows: the English {\sc pos}-bal bands contain 198 nouns, 45 verbs, 64 adjectives and 7 adverbs; similarly for the other two languages.  Greek is not included in this {\sc pos}-based analysis because all sense-annotated  Greek words in EuroSense are nouns. In {\sc freq}-bal, each English band contains 40 words that occur less than 7.1M times in Google Ngrams, and so on and so forth.

 \begin{table}[t]
     \centering
     \scalebox{0.9}{
     \begin{tabular}{c|c|c|c|c}
          \multicolumn{5}{c}{\parbox{7cm}{\centering \vspace{4mm}  {\sc pos}-bal \vspace{2mm}}
          }\\  \hline
          \hline
           & {\bf Nouns} & {\bf Verbs} & {\bf Adjectives} & {\bf Adverbs} \\
           \hline
           en & 198 & 45 & 64 & 7 \\ \cdashline{1-5}
           fr & 171 & 32 & 29 & 9 \\ \cdashline{1-5}
           es & 167 & 22 & 40 & 0 \\ \cdashline{1-5}
           
           \toprule
    \multicolumn{5}{c}{\parbox{7cm}{\centering \vspace{4mm}  {\sc freq}-bal \vspace{2mm}} 
      } \\
     \hline  \hline
         \parbox{4mm}{\multirow{2}{*}{en}} 
          & {\bf 7.1$M$} & {\bf 20$M$} & {\bf 49$M$} & {\bf 682$M$} \\ \cdashline{2-5}
         & 40 & 99 & 62 & 39 \\
          \hline
         \parbox{4mm}{ \multirow{2}{*}{fr} }
           & {\bf 23$m$} & {\bf 70$m$} & {\bf 210$m$} & {\bf 41$M$} \\ \cdashline{2-5}
           & 17 & 43 & 67 & 38 \\
          \hline
         \parbox{4mm}{  \multirow{2}{*}{es} }
            &  {\bf 64$m$} & {\bf 233$m$} & {\bf 793$m$} & {\bf 59$M$} \\ \cdashline{2-5}
           & 12 & 39 & 58 & 48 \\
          \hline
          \parbox{4mm}{ \multirow{2}{*}{el} }
            & {\bf 14$m$} & {\bf 40$m$} & {\bf 111$m$} & {\bf 1.9$M$} \\ \cdashline{2-5}
           & 13 & 41 & 70 & 42 \\
           \toprule
     \end{tabular}     }
     \caption{Content of the polysemy bands in the {\sc pos}-bal and {\sc freq}-bal settings. 
    All bands for a language contain the same number of words of a specific grammatical category or frequency range. $M$ stands for a million and $m$ for a thousand occurrences of a word in a corpus.}
     \label{tab:posfreq_dist_afterbalancing}
 \end{table}

We examine the average $SelfSim$ values obtained for words in each band in {\tt poly-rand}.  Figure \ref{fig:posfreqbalanced_all_langs_mono} shows the  results for monolingual models.  We observe that the {\tt mono} and {\tt poly} words in the {\sc pos}-bal and {\sc freq}-bal bands are ranked similarly to Figure \ref{fig:multi_polysemy_plots}. This shows that BERT's polysemy predictions do not rely on frequency or part of speech. The only exception is Greek BERT  which cannot establish correct  inter-band distinctions when the influence of frequency is neutralised in the {\sc freq}-bal setting. A general observation that applies to all models is that although inter-band distinctions become less clear, the ordering of the bands is preserved. We observe the same trend with ELMo and context2vec.

Statistical tests show that all  inter-band distinctions established by English BERT are still significant in most layers of the model.\footnote{Note that the sample size in this analysis is smaller compared to that used in Sections  \ref{sec:polysemy_results} and \ref{sec:selfsim_based_ranking}.}
This is not the case for ELMo and context2vec, which can distinguish between {\tt mono} and {\tt poly} words but fail to establish significant distinctions between polysemy bands in the balanced settings. For French and Spanish, the statistical analysis shows that all distinctions in {\sc pos}-bal are significant in at least one layer of the models. The same applies to the {\tt mono}$\rightarrow${\tt poly} distinction in {\sc freq}-bal but finer-grained distinctions disappear.\footnote{With a few exceptions: {\tt mono}$\rightarrow${\tt low} and {\tt mid}$\rightarrow${\tt high} are significant in all BETO layers.}

\section{Classification by Polysemy Level} \label{sec:classification}

Our finding that word instance similarity differs across polysemy bands suggests that this feature can be useful for classification. In this Section, we probe the representations for polysemy using a classification experiment where we test their ability to guess whether a word is polysemous, and which {\tt poly} band it falls in. 
We use the {\tt poly-rand} sentence pools and a standard train/dev/test split (70/15/15\%) of the data. 
For the {\tt mono}/{\tt poly} distinction (i.e. the data used in Section \ref{sec:polysemy_identification}), 
this results in 584/126/126 words per set in each language. To guarantee a fair evaluation, we make sure there is no overlap between the 
lemmas in the three sets. 
We use two types of features: (i) the average $SelfSim$ for a word; (ii) all pairwise cosine similarities collected for its instances, which results in 45 features per word ($pairCos$). We train a binary logistic regression classifier for each type of representation and feature.  

As explained in Section \ref{sec:impactlevelpoly}, the three {\tt poly} bands ({\tt low}, {\tt mid} and {\tt high}) and {\tt mono} contain a different number of words. 
For classification into polysemy bands, we balance each class by randomly subsampling words from each band. In total, we use 1,168 words for training, 252 for development and 252 for testing (70/15/15\%) in English. 
In the other languages, we use a split of 840/180/180 words. We train multi-class logistic regression classifiers with the two types of features, $SelfSim$ and $pairCos$.  
We compare the results of the classifiers to a baseline that predicts always the same class, and to a frequency-based classifier which only uses the words' log frequency in Google Ngrams, or in the OSCAR corpus, as a feature. 

Table \ref{tab:classification_results} presents classification accuracy on the test set. We report results obtained with the best layer for each representation type and feature as determined on the development sets. In English, best accuracy is obtained by BERT in both the binary (0.79) and multiclass settings (0.49), followed by mBERT (0.77 and 0.46).
Despite its simplicity, the frequency-based classifier obtains better results than context2vec and ELMo, and performs on par with mBERT in the binary setting. 
This shows that frequency information is highly relevant for the {\tt mono}-{\tt poly} distinction. All classifiers outperform the same class baseline. These results are very encouraging, showing that BERT embeddings can be used to determine whether a word has multiple meanings,  and provide a rough indication of its polysemy level. Results in the other three languages are not as high as those obtained in English, but most models give higher results than the frequency-based classifier.\footnote{Only exceptions are Greek mBERT in the multi-class setting, and Flaubert in both settings.}

\begin{table}[t!]
\centering
\scalebox{0.85}{
\begin{tabular}{p{0.1mm}p{1.7cm}|P{1.15cm}P{1.15cm}|P{1.15cm}P{1.15cm}}
 & & \multicolumn{2}{c|}{{\tt mono/poly}} & \multicolumn{2}{c}{{\tt poly} bands} \\
\hline
& Model & \small{$SelfSim$} & \small{$pairCos$} & \small{$SelfSim$} & \small{$pairCos$}\\ 
\hline
\multirow{5}{*}{\rotatebox[origin=c]{90}{\sc en}} & BERT &  0.76$_{10}$ & \textbf{0.79}$_{8}$   &  \textbf{0.49}$_{10}$ &  0.46$_{10}$ \\
& mBERT & 0.77$_{8}$ & 0.75$_{8}$ & 0.46$_{12}$ & 0.43$_{12}$ \\
& ELMo &  0.69$_{2}$ & 0.63$_{3}$ &  0.37$_{2}$  &  0.34$_{3}$ \\
& context2vec &  0.61  & 0.61&   0.34 &  0.31 \\

\cdashline{2-6} 
& Frequency & \multicolumn{2}{c|}{0.77} & \multicolumn{2}{c}{0.41} \\
\hline
\parbox[t]{0.1mm}{\multirow{3}{*}{\rotatebox[origin=c]{90}{\sc fr}}} & Flaubert & 0.58$_{7}$  & 0.55$_{6}$ & 0.29$_{8}$ & 0.27$_{9}$ \\
& mBERT &  \textbf{0.66}$_{9}$ &  0.64$_{9}$ &  \textbf{0.38}$_{7}$ &  \textbf{0.38}$_{8}$\\
\cdashline{2-6}
& Frequency & \multicolumn{2}{c|}{0.61}  & \multicolumn{2}{c}{0.37} \\
\hline
\parbox[t]{0.1mm}{\multirow{3}{*}{\rotatebox[origin=c]{90}{\sc es}}} & BETO & \textbf{0.70}$_{9}$ & 0.66$_{7}$ & 0.42$_{6}$ & \textbf{0.48}$_{5}$ \\
& mBERT & 0.69$_{11}$ & 0.64$_{7}$ & 0.38$_{9}$ & 0.43$_{7}$\\

\cdashline{2-6} 
& Frequency & \multicolumn{2}{c|}{0.67} & \multicolumn{2}{c}{0.41}\\

\hline
\parbox[t]{0.1mm}{\multirow{3}{*}{\rotatebox[origin=c]{90}{\sc el}}} & \small{GreekBERT} & \textbf{0.70}$_{4}$ & 0.64$_{4}$ & 0.34$_{4}$ & \textbf{0.38}$_{6}$ \\
& mBERT & 0.60$_{7}$  & 0.65$_{7}$ &0.32$_{11}$ & 0.34$_{9}$\\
\cdashline{2-6} 
& Frequency & \multicolumn{2}{c|}{0.63} & \multicolumn{2}{c}{0.35} \\
\hline
& Baseline & \multicolumn{2}{c|}{0.50} & \multicolumn{2}{c}{0.25} \\

\end{tabular}}
\caption{Accuracy of binary ({\tt mono/poly}) and multi-class ({\tt poly} bands) classifiers using $SelfSim$ and $pairCos$ features on the test sets. Comparison to a baseline that predicts always the same class and a classifier that only uses log frequency as feature. Subscripts denote the layers used.} 
\label{tab:classification_results}
\end{table}

\section{Word Sense Clusterability} \label{sec:clusterability}

We have shown that representations from pre-trained LMs encode rich information about words' degree of polysemy. They can successfully distinguish {\tt mono} from {\tt poly} lemmas, and predict the polysemy level of words. Our previous experiments involved a set of controlled settings representing different sense distributions and polysemy levels. 
In this Section, we explore whether these representations can also point to the  clusterability of {\tt poly} words in an uncontrolled setting. 

\subsection{Task Definition}

Instances of some {\tt poly} words are  easier to group into interpretable clusters than others. This is, for example, a simple task for the ambiguous noun {\it rock} which can express two clearly separate senses ({\sc stone} and {\sc music}), but harder for {\it book} which might refer to the {\sc content} or {\sc object} senses of the word (e.g. {\it I read a \underline{book}} vs. {\it I bought a \underline{book}}). 
In what follows, we test the ability of contextualised  representations to estimate how easy this task is for a specific word, i.e. its partitionability into senses.

Following \citet{mccarthy2016word}, we use the clusterability metrics proposed by \citet{ackerman2009clusterability} to measure the ease of clustering word instances into senses. McCarthy et al.  base their clustering on the similarity of manual meaning-preserving annotations (lexical substitutes  and translations). Instances of different senses, such as: {\it Put granola \underline{bars} in a bowl} vs. {\it That's not a very high \underline{bar}}, present no overlap in their in-context substitutes: \{{\it snack}, {\it biscuit}, {\it block}, {\it slab}\} vs. \{{\it pole}, {\it marker}, {\it hurdle}, {\it barrier}, {\it level}, {\it obstruction}\}. Semantically related instances, on the contrary, share a different number of substitutes depending on their proximity. 
The need for manual annotations, however, constrains the method's applicability to specific datasets.

We propose to extend and scale up the \citet{mccarthy2016word} clusterability approach using contextualised representations, in order to  make it applicable to a larger vocabulary. These experiments are carried out in English due to the lack of evaluation data in other languages. 

\subsection{Data}

We run our experiments on the usage similarity (Usim) dataset \cite{erk2013measuring} for comparison with previous work. 
Usim contains ten instances for 56 target words of different PoS from the SemEval  Lexical Substitution dataset \cite{mccarthy2007semeval}.  
Word instances are manually annotated with pairwise similarity scores on a scale from 1 (completely different) to 5 (same meaning).

We represent target word instances in Usim in two ways: using {\bf contextualised representations} generated by BERT, context2vec and ELMo (BERT-{\sc rep}, c2v-{\sc rep}, ELMo-{\sc rep});\footnote{We do not use the first layer of ELMo in this experiment. It is character-based, so most representations of a lemma are identical and we cannot obtain meaningful clusters.} using {\bf substitute-based representations} 
with  automatically generated substitutes. The substitute-based approach allows for a direct comparison with the method of \citet{mccarthy2016word}. 
They represent each instance $i$ of a word $w$ in Usim   as a vector $\vec{i}$, where each substitute $s$  assigned to $w$  over  all its instances ($i$ $\in$  $I $)  becomes a dimension ($d_{s}$). For a given $i$, the value for each  $d_{s}$ is the number of annotators who proposed  substitute $s$. $d_{s}$ contains a zero entry if $s$ was not proposed for $i$. 
We refer to this type of representation as {\tt Gold-{\sc sub}}. 
We generate our substitute-based representations with BERT using the simple ``word similarity''  approach in \citet{zhou-etal-2019-bert}. For an instance $i$ of word $w$ in context $C$, we rank a set of candidate substitutes $S = \{s_1, s_2, ..., s_n$\} based on the cosine similarity of the BERT representations for $i$ and for each substitute $s_{j}$ $\in S$ in the same context $C$. We use representations from the last layer of the model. As candidate substitutes, we use the unigram paraphrases of $w$ in the Paraphrase Database (PPDB) XXL package \cite{ganitkevitch2013ppdb,pavlick2015ppdb}.\footnote{We use PPDB (\url{http://www.paraphrase.org}) to reduce variability in our substitute sets, compared to the ones that would be proposed by looking at the whole vocabulary.} 

For each instance $i$ of $w$, we obtain a ranking $R$ of all substitutes in $S$. We remove low-quality substitutes (i.e. noisy paraphrases or substitutes referring to a different sense of $w$) using the  filtering approach proposed by \citet{gari-soler-etal-2019-word}. 
We check each pair of substitutes in subsequent positions in $R$, starting from the top; if a pair is unrelated in PPDB, all substitutes from that position onwards are discarded.  
The idea is that good quality substitutes should be both high-ranked and semantically related.  
We build vectors as in \citet{mccarthy2016word}, using the cosine similarity assigned by BERT to each substitute as a value. We call this representation BERT-{\sc sub}.

\begin{table*}[t]
\centering
\scalebox{0.87}{
\begin{tabular}{p{1.2cm}|c||ccc|cc|cc}

   \parbox{1.2cm}{\begin{center} Gold  \end{center}}
  & Metric & BERT-{\sc rep} &  c2v-{\sc rep} & ELMo-{\sc rep} & BERT-{\sc sub} & {\tt Gold}-{\sc sub}  & \sc{BERT-Agg} &  {\tt Gold}-{\sc Agg}  \\
 \hline

\multirow{3}{*}{\parbox{1.2cm}{\begin{center} Uiaa \end{center}}} & {\sc sep} $\searrow$  & 
-0.48*$_{10}$ & -0.12 &  -0.24$_{2}$ &   -0.03  & -0.20  & -0.48*$_{11}$ & -- \\

& {\sc vr} $\nearrow$
& 0.17$_{12}$ &    0.14  & 0.19$_{2}$  & 0.09 & 0.34* & 0.33*$_{12}$ & -- \\ 

& {\sc sil} $\nearrow$
& 0.61*$_{11}$ &  0.06  &   0.21$_{2}$ &  0.10 & 0.32* & \textbf{0.69}*$_{10}$ & \textbf{0.80}* \\
\hline

\multirow{3}{*}{\parbox{1.2cm}{\begin{center} Umid \end{center}}} & {\sc sep} $\nearrow$ & 
0.43*$_9$  & -0.01  &  0.08$_{3}$ &  0.05 & 0.16 & 0.43*$_9$ & -- \\ 
 
& {\sc vr} $\searrow$ &  
-0.24$_9$ & -0.08 &  -0.15$_{3}$  &-0.15 & -0.24  & -0.32*$_5$ & -- \\ 
& {\sc sil} $\searrow$
& \textbf{-0.46}*$_{10}$ & 0.05 & -0.06$_{2}$ &  -0.11 & -0.38* & -0.44*$_8$ & \textbf{-0.48}* \\

\end{tabular}
}
\caption{Spearman's $\rho$ correlation between automatic metrics and gold standard clusterability estimates. Significant correlations (where the null hypothesis $\rho = 0$ is rejected with $\alpha$ $<$ 0.05) are marked with *. The arrows indicate the expected direction of correlation for each metric. Subscripts indicate the layer that achieved best performance. 
The two strongest correlations obtained with each gold standard measure are in boldface.} 
\label{tab:results_umiduiaa}
\end{table*}

\subsection{Sense Clustering}

The clusterability metrics that we use are metrics initially proposed for estimating the quality of the optimal clustering that can be obtained from a dataset; the better the quality of this clustering, the higher the clusterability of the dataset it is derived from \citep{ackerman2009clusterability}.

In order to estimate the clusterability of a word $w$, we thus need to first cluster its instances in the data. We use the $k$-means algorithm which requires the number of senses for a lemma. This is, of course, different for every lemma in our dataset.
We define the optimal number of clusters $k$ for a lemma in a data-driven manner using the Silhouette coefficient ({\sc sil}) \cite{rousseeuw1987silhouettes}, without recourse to external resources.\footnote{We do not use McCarthy et al.'s graph-based approach because it is not compatible with all our representation types.} For a data point $i$, {\sc sil} compares the intra-cluster distance 
(i.e. the average distance from $i$ to every other data point in the same cluster) with the average distance of $i$ to all points in its nearest cluster. 
The {\sc sil} value for a clustering is obtained by averaging {\sc sil} for all data points, and it ranges from -1 to 1. 
We cluster each type of representation for $w$ 
using $k$-means with a range of $k$ values ($2 \leq k \leq 10$), and retain the $k$ of the clustering with the highest mean {\sc sil}. 
Additionally, since BERT representations' cosine similarity correlates well with usage similarity \cite{gari-soler-etal-2019-word}, we experiment with Agglomerative Clustering with average linkage directly on the cosine distance matrix obtained with BERT representations ({\sc BERT-Agg}). For comparison, we also use Agglomerative Clustering on the
gold usage similarity scores from Usim, transformed into distances ({\tt Gold-{\sc Agg}}).

\subsection{Clusterability Metrics} \label{sec:measures}

We use in our experiments the two best performing  
metrics from McCarthy et al. \shortcite{mccarthy2016word}:  
Variance Ratio (\textsc{vr}) \cite{zhang2001dependence}  and Separability ({\sc sep}) \cite{ostrovsky2012effectiveness}. 
\textsc{vr}  calculates the ratio of the within- and  between-cluster variance for a given clustering solution. {\sc sep} measures the difference in loss between two clusterings with $k - 1$ and $k$ clusters and its range is [0,1). We use $k$-means' sum of squared distances of data points to their closest cluster center as the loss. 
Details about these two metrics are given in Appendix \ref{app:metrics}.\footnote{Note that the {\sc vr} and {\sc sep} metrics are not compatible with {\tt Gold-{\sc Agg}}  which relies on Usim similarity scores, because we need vectors for their calculation. For BERT-{\sc Agg}, we calculate {\sc vr} and {\sc sep} using BERT embeddings.} We also experiment with  \textsc{sil} as a clusterability metric, as it can assess cluster validity. For {\sc vr} and {\sc sil}, a higher value indicates higher clusterability. The inverse applies to {\sc sep}. 

We calculate Spearman's $\rho$ correlation between the results of each clusterability metric and two gold standard measures derived from Usim: {\bf Uiaa} and {\bf Umid}. 
Uiaa is the inter-annotator agreement for a lemma in terms of average pairwise Spearman's correlation between annotators' judgments. 
Higher Uiaa values indicate higher clusterability, meaning that sense partitions are clearer and easier to agree upon. 
Umid is the proportion of mid-range judgments (between 2 and 4) assigned by annotators to all instances of a target word. It indicates how often usages {\underline {do not}} have identical (5) or completely different (1) meaning. Therefore, higher Umid values indicate lower clusterability. 

\subsection{Results and Discussion}

The clusterability results are given in Table \ref{tab:results_umiduiaa}.  Agglomerative Clustering 
on the gold Usim similarity scores ({\tt Gold-{\sc Agg}}) gives best results on the Uiaa evaluation in combination with the {\sc sil} clusterability metric ($\rho =
0.80$). This is unsurprising, since Uiaa and Umid are derived from the same Usim scores.
From our automatically generated representations, the strongest correlation with Uiaa (0.69) is obtained with {\sc BERT-Agg} and the {\sc sil} clusterability metric. 
The {\sc sil} metric also works well with BERT-{\sc rep} achieving the strongest correlation with Umid (-0.46). 
It constitutes, thus, a good alternative to the {\sc sep} and {\sc vr} metrics used in previous studies when combined with BERT representations. 

Interestingly, the correlations obtained using raw BERT contextualised representations are much higher than the ones 
observed with \citet{mccarthy2016word}'s representations which rely on manual substitutes ({\tt Gold}-{\sc sub}). These were in the range of 0.20-0.34  for Uiaa and 0.16-0.38 for Umid (in absolute value). 
The results demonstrate that \textbf{BERT representations offer good estimates of the partitionability of words into senses}, improving over substitute annotations. 
As expected, the substitution-based approach performs better  with clean manual substitutes ({\tt Gold-{\sc sub}}) than with automatically generated ones 
(BERT-{\sc sub}).

\begin{figure}
    \centering
    \includegraphics[width=.7\columnwidth]{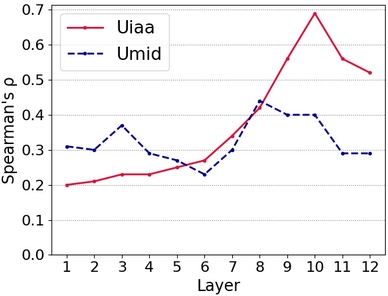}
    \caption{Spearman's $\rho$ correlations 
    between the gold standard Uiaa and Umid 
    scores, and clusterability estimates obtained using Agglomerative Clustering on a cosine distance matrix of BERT representations. 
}
    \label{fig:bertclus_bylayer}
\end{figure}

We present a per layer analysis of the correlations obtained with the best performing BERT representations (BERT-{\sc Agg}) and the  {\sc sil}  metric in Figure \ref{fig:bertclus_bylayer}. 
We report the absolute values of the correlation coefficient for a more straightforward comparison. For Uiaa, the higher layers of the model make the best predictions: correlations increase monotonically up to layer 10, and then they show a slight decrease. Umid prediction shows a more irregular pattern: it peaks at layers 3 and 8, and decreases again in the last layers.

\section{Conclusion}

We have shown that contextualised BERT representations encode rich information about lexical polysemy. Our experimental results suggest that this high quality knowledge about words, which allows BERT to detect polysemy in different configurations and across all layers, is  acquired during pre-training. 
Our findings hold for the English BERT as well as for BERT models in other languages, as shown by our experiments on French, Spanish and Greek, and to a lesser extent for multilingual BERT. 
Moreover, English BERT representations can be used to obtain a good estimation of a word's partitionability into senses. These results open up new avenues for research in multilingual semantic analysis, and we can consider various theoretical and application-related extensions for this work.  

The polysemy and sense-related knowledge revealed by the models can serve to develop novel methodologies for improved cross-lingual alignment of embedding spaces and cross-lingual transfer, pointing to more polysemous (or less clusterable) words for which transfer might be harder. Predicting the polysemy level of words  can also be useful for determining the context needed for acquiring representations that properly reflect the meaning of word instances in running text. From a more theoretical standpoint, we expect this work to be useful for studies on the organisation of the semantic space in different languages and on lexical semantic change.

\section*{Acknowledgements}

\setlength\intextsep{0mm}

\begin{wrapfigure}{L}{0pt}
\includegraphics[scale=0.3]{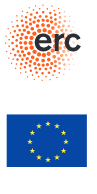}
\end{wrapfigure}

This work has been supported by the French National Research Agency under project ANR-16-CE33-0013. The work is also part of the FoTran project, funded by the European Research Council (ERC) under the European Union’s Horizon 2020 research and innovation programme (grant agreement \textnumero ~771113). We thank the anonymous reviewers and the TACL Action Editor for their careful reading of our paper, their thorough reviews and their helpful suggestions.

\bibliography{tacl2018}

\begin{thebibliography}{67}
\expandafter\ifx\csname natexlab\endcsname\relax\def\natexlab#1{#1}\fi

\bibitem[{Ackerman and Ben-David(2009)}]{ackerman2009clusterability}
Margareta Ackerman and Shai Ben-David. 2009.
\newblock \href {http://proceedings.mlr.press/v5/ackerman09a.html}
  {{Clusterability: A Theoretical Study}}.
\newblock \emph{Journal of Machine Learning Research}, 5:1--8.

\bibitem[{Adi et~al.(2017)Adi, Kermany, Belinkov, Lavi, and
  Goldberg}]{Adi2016FinegrainedAO}
Yossi Adi, Einat Kermany, Yonatan Belinkov, Ofer Lavi, and Yoav Goldberg. 2017.
\newblock \href {https://openreview.net/pdf?id=BJh6Ztuxl} {{Fine-grained
  Analysis of Sentence Embeddings Using Auxiliary Prediction Tasks}}.
\newblock In \emph{Proceedings of ICLR}, Toulon, France.

\bibitem[{Agirre and Martinez(2004)}]{agirre-martinez-2004-unsupervised}
Eneko Agirre and David Martinez. 2004.
\newblock \href {https://www.aclweb.org/anthology/W04-3204} {{Unsupervised
  {WSD} based on Automatically Retrieved Examples: The Importance of Bias}}.
\newblock In \emph{Proceedings of the 2004 Conference on Empirical Methods in
  Natural Language Processing}, pages 25--32, Barcelona, Spain. Association for
  Computational Linguistics.

\bibitem[{Aina et~al.(2019)Aina, Gulordava, and
  Boleda}]{aina-etal-2019-putting}
Laura Aina, Kristina Gulordava, and Gemma Boleda. 2019.
\newblock \href {https://doi.org/10.18653/v1/P19-1324} {{Putting Words in
  Context: {LSTM} Language Models and Lexical Ambiguity}}.
\newblock In \emph{Proceedings of the 57th Annual Meeting of the Association
  for Computational Linguistics}, pages 3342--3348, Florence, Italy.
  Association for Computational Linguistics.

\bibitem[{Baroni et~al.(2009)Baroni, Bernardini, Ferraresi, and
  Zanchetta}]{baroni2009wacky}
Marco Baroni, Silvia Bernardini, Adriano Ferraresi, and Eros Zanchetta. 2009.
\newblock \href
  {https://wacky.sslmit.unibo.it/lib/exe/fetch.php?media=papers:wacky_2008.pdf}
  {{The WaCky Wide Web: a Collection of Very Large Linguistically Processed
  Web-Crawled Corpora}}.
\newblock \emph{Journal of Language Resources and Evaluation}, 43(3):209--226.

\bibitem[{Benjamini and Hochberg(1995)}]{benjamini1995controlling}
Yoav Benjamini and Yosef Hochberg. 1995.
\newblock \href
  {https://rss.onlinelibrary.wiley.com/doi/10.1111/j.2517-6161.1995.tb02031.x}
  {{Controlling the False Discovery Rate: a Practical and Powerful Approach to
  Multiple Testing}}.
\newblock \emph{Journal of the Royal statistical society: series B
  (Methodological)}, 57(1):289--300.

\bibitem[{Bird et~al.(2009)Bird, Klein, and Loper}]{bird2009natural}
Steven Bird, Ewan Klein, and Edward Loper. 2009.
\newblock \href {https://www.nltk.org/book/} {\emph{{Natural Language
  Processing with Python: Analyzing Text with the Natural Language Toolkit}}}.
\newblock O'Reilly Media, Inc., Beijing.

\bibitem[{Brants and Franz(2006)}]{brants2006web}
Thorsten Brants and Alex Franz. 2006.
\newblock \href {https://catalog.ldc.upenn.edu/LDC2006T13} {{Web 1T 5-gram
  Version 1}}.
\newblock In \emph{LDC2006T13}, Philadelphia, Pennsylvania. Linguistic Data
  Consortium.

\bibitem[{Camacho-Collados and
  Pilehvar(2018)}]{journals/jair/Camacho-Collados18}
Jos\'e Camacho-Collados and Mohammad~Taher Pilehvar. 2018.
\newblock \href {https://www.jair.org/index.php/jair/article/view/11259/26454}
  {{From Word To Sense Embeddings: A Survey on Vector Representations of
  Meaning}}.
\newblock \emph{Journal of Artificial Intelligence Research}, 63:743--788.

\bibitem[{Cañete et~al.(2020)Cañete, Chaperon, Fuentes, and
  Pérez}]{CaneteCFP2020}
José Cañete, Gabriel Chaperon, Rodrigo Fuentes, and Jorge Pérez. 2020.
\newblock \href {https://users.dcc.uchile.cl/~jperez/papers/pml4dc2020.pdf}
  {{Spanish Pre-Trained BERT Model and Evaluation Data}}.
\newblock In \emph{Proceedings of the ICLR 2020 Workshop on Practical ML for
  Developing Countries (PML4DC)}, Addis Ababa, Ethiopia.

\bibitem[{Clark et~al.(2019)Clark, Khandelwal, Levy, and
  Manning}]{clark-etal-2019-bert}
Kevin Clark, Urvashi Khandelwal, Omer Levy, and Christopher~D. Manning. 2019.
\newblock \href {https://doi.org/10.18653/v1/W19-4828} {{What Does {BERT} Look
  at? An Analysis of {BERT}{'}s Attention}}.
\newblock In \emph{Proceedings of the ACL 2019 Workshop BlackboxNLP: Analyzing
  and Interpreting Neural Networks for NLP}, pages 276--286, Florence, Italy.
  Association for Computational Linguistics.

\bibitem[{Conneau et~al.(2018)Conneau, Kruszewski, Lample, Barrault, and
  Baroni}]{conneau-etal-2018-cram}
Alexis Conneau, German Kruszewski, Guillaume Lample, Lo{\"\i}c Barrault, and
  Marco Baroni. 2018.
\newblock \href {https://doi.org/10.18653/v1/P18-1198} {What you can cram into
  a single {\$}{\&}!{\#}* vector: Probing sentence embeddings for linguistic
  properties}.
\newblock In \emph{Proceedings of the 56th Annual Meeting of the Association
  for Computational Linguistics (Volume 1: Long Papers)}, pages 2126--2136,
  Melbourne, Australia. Association for Computational Linguistics.

\bibitem[{Delli~Bovi et~al.(2017)Delli~Bovi, Camacho-Collados, Raganato, and
  Navigli}]{bovi2017eurosense}
Claudio Delli~Bovi, Jose Camacho-Collados, Alessandro Raganato, and Roberto
  Navigli. 2017.
\newblock \href {https://doi.org/10.18653/v1/P17-2094} {{EuroSense: Automatic
  Harvesting of Multilingual Sense Annotations from Parallel Text}}.
\newblock In \emph{Proceedings of the 55th Annual Meeting of the Association
  for Computational Linguistics (Volume 2: Short Papers)}, pages 594--600,
  Vancouver, Canada. Association for Computational Linguistics.

\bibitem[{Devlin et~al.(2019)Devlin, Chang, Lee, and
  Toutanova}]{devlin2019bert}
Jacob Devlin, Ming-Wei Chang, Kenton Lee, and Kristina Toutanova. 2019.
\newblock \href {https://doi.org/10.18653/v1/N19-1423} {{BERT: Pre-training of
  Deep Bidirectional Transformers for Language Understanding}}.
\newblock In \emph{Proceedings of the 2019 Conference of the North {A}merican
  Chapter of the Association for Computational Linguistics: Human Language
  Technologies, Volume 1 (Long and Short Papers)}, pages 4171--4186,
  Minneapolis, Minnesota. Association for Computational Linguistics.

\bibitem[{Dubossarsky et~al.(2019)Dubossarsky, Hengchen, Tahmasebi, and
  Schlechtweg}]{dubossarsky-etal-2019-time}
Haim Dubossarsky, Simon Hengchen, Nina Tahmasebi, and Dominik Schlechtweg.
  2019.
\newblock \href {https://doi.org/10.18653/v1/P19-1044} {{Time-Out: Temporal
  Referencing for Robust Modeling of Lexical Semantic Change}}.
\newblock In \emph{Proceedings of the 57th Annual Meeting of the Association
  for Computational Linguistics}, pages 457--470, Florence, Italy. Association
  for Computational Linguistics.

\bibitem[{Erk et~al.(2009)Erk, McCarthy, and Gaylord}]{erketal2009}
Katrin Erk, Diana McCarthy, and Nicholas Gaylord. 2009.
\newblock \href {http://aclweb.org/anthology/P09-1002} {{Investigations on Word
  Senses and Word Usages}}.
\newblock In \emph{Proceedings of the Joint Conference of the 47th Annual
  Meeting of the ACL and the 4th International Joint Conference on Natural
  Language Processing of the AFNLP}, pages 10--18, Suntec, Singapore.
  Association for Computational Linguistics.

\bibitem[{Erk et~al.(2013)Erk, McCarthy, and Gaylord}]{erk2013measuring}
Katrin Erk, Diana McCarthy, and Nicholas Gaylord. 2013.
\newblock \href {https://doi.org/10.1162/COLI_a_00142} {{Measuring Word Meaning
  in Context}}.
\newblock \emph{Computational Linguistics}, 39(3):511--554.

\bibitem[{Ethayarajh(2019)}]{ethayarajh-2019-contextual}
Kawin Ethayarajh. 2019.
\newblock \href {https://doi.org/10.18653/v1/D19-1006} {{How Contextual are
  Contextualized Word Representations? Comparing the Geometry of {BERT},
  {ELM}o, and {GPT}-2 Embeddings}}.
\newblock In \emph{Proceedings of the 2019 Conference on Empirical Methods in
  Natural Language Processing and the 9th International Joint Conference on
  Natural Language Processing (EMNLP-IJCNLP)}, pages 55--65, Hong Kong, China.
  Association for Computational Linguistics.

\bibitem[{Fellbaum(1998)}]{Fellbaum1998}
Christiane Fellbaum. 1998.
\newblock \href {http://mitpress.mit.edu/books/wordnet} {\emph{{WordNet: An
  Electronic Lexical Database}}}.
\newblock Language, Speech, and Communication. MIT Press, Cambridge, MA.

\bibitem[{Ganitkevitch et~al.(2013)Ganitkevitch, Van~Durme, and
  Callison-Burch}]{ganitkevitch2013ppdb}
Juri Ganitkevitch, Benjamin Van~Durme, and Chris Callison-Burch. 2013.
\newblock \href {http://aclweb.org/anthology/N13-1092} {{PPDB: The Paraphrase
  Database}}.
\newblock In \emph{Proceedings of the 2013 Conference of the North American
  Chapter of the Association for Computational Linguistics: Human Language
  Technologies}, pages 758--764, Atlanta, Georgia. Association for
  Computational Linguistics.

\bibitem[{Gar{\'\i}~Soler et~al.(2019)Gar{\'\i}~Soler, Apidianaki, and
  Allauzen}]{gari-soler-etal-2019-word}
Aina Gar{\'\i}~Soler, Marianna Apidianaki, and Alexandre Allauzen. 2019.
\newblock \href {https://doi.org/10.18653/v1/S19-1002} {{Word Usage Similarity
  Estimation with Sentence Representations and Automatic Substitutes}}.
\newblock In \emph{Proceedings of the Eighth Joint Conference on Lexical and
  Computational Semantics (*{SEM} 2019)}, pages 9--21, Minneapolis, Minnesota.
  Association for Computational Linguistics.

\bibitem[{Giulianelli et~al.(2020)Giulianelli, Del~Tredici, and
  Fern{\'a}ndez}]{giulianelli-etal-2020-analysing}
Mario Giulianelli, Marco Del~Tredici, and Raquel Fern{\'a}ndez. 2020.
\newblock \href {https://doi.org/10.18653/v1/2020.acl-main.365} {{Analysing
  Lexical Semantic Change with Contextualised Word Representations}}.
\newblock In \emph{Proceedings of the 58th Annual Meeting of the Association
  for Computational Linguistics}, pages 3960--3973, Online. Association for
  Computational Linguistics.

\bibitem[{Hewitt and Liang(2019)}]{hewitt-liang-2019-designing}
John Hewitt and Percy Liang. 2019.
\newblock \href {https://doi.org/10.18653/v1/D19-1275} {{Designing and
  Interpreting Probes with Control Tasks}}.
\newblock In \emph{Proceedings of the 2019 Conference on Empirical Methods in
  Natural Language Processing and the 9th International Joint Conference on
  Natural Language Processing (EMNLP-IJCNLP)}, pages 2733--2743, Hong Kong,
  China. Association for Computational Linguistics.

\bibitem[{Hewitt and Manning(2019)}]{hewitt2019structural}
John Hewitt and Christopher~D. Manning. 2019.
\newblock \href {https://doi.org/10.18653/v1/N19-1419} {{A Structural Probe for
  Finding Syntax in Word Representations}}.
\newblock In \emph{Proceedings of the 2019 Conference of the North {A}merican
  Chapter of the Association for Computational Linguistics: Human Language
  Technologies, Volume 1 (Long and Short Papers)}, pages 4129--4138,
  Minneapolis, Minnesota. Association for Computational Linguistics.

\bibitem[{Huang et~al.(2012)Huang, Socher, Manning, and
  Ng}]{Huang:2012:IWR:2390524.2390645}
Eric Huang, Richard Socher, Christopher Manning, and Andrew Ng. 2012.
\newblock \href {https://www.aclweb.org/anthology/P12-1092} {{Improving Word
  Representations via Global Context and Multiple Word Prototypes}}.
\newblock In \emph{Proceedings of the 50th Annual Meeting of the Association
  for Computational Linguistics (Volume 1: Long Papers)}, pages 873--882, Jeju
  Island, Korea. Association for Computational Linguistics.

\bibitem[{Jakubowski et~al.(2020)Jakubowski, Gasic, and
  Zibrowius}]{jakubowski-etal-2020-topology}
Alexander Jakubowski, Milica Gasic, and Marcus Zibrowius. 2020.
\newblock \href {https://www.aclweb.org/anthology/2020.starsem-1.11} {{Topology
  of Word Embeddings: Singularities Reflect Polysemy}}.
\newblock In \emph{Proceedings of the Ninth Joint Conference on Lexical and
  Computational Semantics}, pages 103--113, Barcelona, Spain (Online).
  Association for Computational Linguistics.

\bibitem[{Kilgarriff(2004)}]{Kilgarriff2004}
Adam Kilgarriff. 2004.
\newblock \href
  {https://link.springer.com/chapter/10.1007%2F978-3-540-30120-2_14} {{How
  Dominant Is the Commonest Sense of a Word?}}
\newblock Lecture Notes in Computer Science (vol. 3206), Text, Speech and
  Dialogue, Sojka Petr, Kopeček Ivan, Pala Karel (eds.), pages 103--112.
  Springer, Berlin, Heidelberg.

\bibitem[{Koutsikakis et~al.(2020)Koutsikakis, Chalkidis, Malakasiotis, and
  Androutsopoulos}]{koutsikakis2020greek}
John Koutsikakis, Ilias Chalkidis, Prodromos Malakasiotis, and Ion
  Androutsopoulos. 2020.
\newblock \href {https://dl.acm.org/doi/10.1145/3411408.3411440} {{GREEK-BERT:
  The Greeks visiting Sesame Street}}.
\newblock In \emph{Proceedings of the 11th Hellenic Conference on Artificial
  Intelligence (SETN 2020)}, pages 110--117, Athens, Greece.

\bibitem[{Kovaleva et~al.(2019)Kovaleva, Romanov, Rogers, and
  Rumshisky}]{kovaleva-etal-2019-revealing}
Olga Kovaleva, Alexey Romanov, Anna Rogers, and Anna Rumshisky. 2019.
\newblock \href {https://doi.org/10.18653/v1/D19-1445} {{Revealing the Dark
  Secrets of {BERT}}}.
\newblock In \emph{Proceedings of the 2019 Conference on Empirical Methods in
  Natural Language Processing and the 9th International Joint Conference on
  Natural Language Processing (EMNLP-IJCNLP)}, pages 4365--4374, Hong Kong,
  China. Association for Computational Linguistics.

\bibitem[{Le et~al.(2020)Le, Vial, Frej, Segonne, Coavoux, Lecouteux, Allauzen,
  Crabb\'{e}, Besacier, and Schwab}]{le2020flaubert}
Hang Le, Lo\"{i}c Vial, Jibril Frej, Vincent Segonne, Maximin Coavoux, Benjamin
  Lecouteux, Alexandre Allauzen, Beno\^{i}t Crabb\'{e}, Laurent Besacier, and
  Didier Schwab. 2020.
\newblock \href {https://www.aclweb.org/anthology/2020.lrec-1.302} {{FlauBERT:
  Unsupervised Language Model Pre-training for French}}.
\newblock In \emph{Proceedings of The 12th Language Resources and Evaluation
  Conference}, pages 2479--2490, Marseille, France. European Language Resources
  Association.

\bibitem[{Leacock et~al.(1998)Leacock, Chodorow, and
  Miller}]{leacock-etal-1998-using}
Claudia Leacock, Martin Chodorow, and George~A. Miller. 1998.
\newblock \href {https://www.aclweb.org/anthology/J98-1006} {{Using Corpus
  Statistics and {W}ord{N}et Relations for Sense Identification}}.
\newblock \emph{Computational Linguistics}, 24(1):147--165.

\bibitem[{Linzen(2018)}]{Linzen2018}
Tal Linzen. 2018.
\newblock \href {https://arxiv.org/abs/1809.04179} {What can linguistics and
  deep learning contribute to each other?}
\newblock \emph{arXiv preprint:1809.04179v2}.

\bibitem[{Linzen et~al.(2016)Linzen, Dupoux, and
  Goldberg}]{linzen2016assessing}
Tal Linzen, Emmanuel Dupoux, and Yoav Goldberg. 2016.
\newblock \href {https://doi.org/10.1162/tacl_a_00115} {{Assessing the Ability
  of {LSTM}s to Learn Syntax-Sensitive Dependencies}}.
\newblock \emph{Transactions of the Association for Computational Linguistics},
  4:521--535.

\bibitem[{Loureiro and Camacho-Collados(2020)}]{camachocollados2020}
Daniel Loureiro and Jose Camacho-Collados. 2020.
\newblock \href {https://doi.org/10.18653/v1/2020.emnlp-main.283} {Don{'}t
  neglect the obvious: On the role of unambiguous words in word sense
  disambiguation}.
\newblock In \emph{Proceedings of the 2020 Conference on Empirical Methods in
  Natural Language Processing (EMNLP)}, pages 3514--3520, Online. Association
  for Computational Linguistics.

\bibitem[{McCarthy et~al.(2016)McCarthy, Apidianaki, and
  Erk}]{mccarthy2016word}
Diana McCarthy, Marianna Apidianaki, and Katrin Erk. 2016.
\newblock \href
  {https://www.mitpressjournals.org/doi/pdfplus/10.1162/COLI_a_00247} {{Word
  Sense Clustering and Clusterability}}.
\newblock \emph{Computational Linguistics}, 42(2):245--275.

\bibitem[{McCarthy et~al.(2004)McCarthy, Koeling, Weeds, and
  Carroll}]{mccarthy-etal-2004-finding}
Diana McCarthy, Rob Koeling, Julie Weeds, and John Carroll. 2004.
\newblock \href {https://doi.org/10.3115/1218955.1218991} {{Finding Predominant
  Word Senses in Untagged Text}}.
\newblock In \emph{Proceedings of the 42nd Annual Meeting of the Association
  for Computational Linguistics ({ACL}-04)}, pages 279--286, Barcelona, Spain.

\bibitem[{McCarthy and Navigli(2007)}]{mccarthy2007semeval}
Diana McCarthy and Roberto Navigli. 2007.
\newblock \href {http://aclweb.org/anthology/S07-1009} {{SemEval-2007 Task 10:
  English Lexical Substitution Task}}.
\newblock In \emph{Proceedings of the Fourth International Workshop on Semantic
  Evaluations (SemEval-2007)}, pages 48--53, Prague, Czech Republic.
  Association for Computational Linguistics.

\bibitem[{Melamud et~al.(2016)Melamud, Goldberger, and
  Dagan}]{melamud2016context2vec}
Oren Melamud, Jacob Goldberger, and Ido Dagan. 2016.
\newblock \href {https://doi.org/10.18653/v1/K16-1006} {{context2vec: Learning
  Generic Context Embedding with Bidirectional LSTM}}.
\newblock In \emph{Proceedings of the 20th SIGNLL Conference on Computational
  Natural Language Learning}, pages 51--61, Berlin, Germany. Association for
  Computational Linguistics.

\bibitem[{Mikolov et~al.(2013)Mikolov, Chen, Corrado, and
  Dean}]{mikolov2013efficient}
Tomas Mikolov, Kai Chen, Greg Corrado, and Jeffrey Dean. 2013.
\newblock \href {https://arxiv.org/abs/1301.3781} {{Efficient Estimation of
  Word Representations in Vector Space}}.
\newblock \emph{arXiv preprint:1301.3781v3}.

\bibitem[{Miller et~al.(1993)Miller, Leacock, Tengi, and
  Bunker}]{miller-etal-1993-semantic}
George~A. Miller, Claudia Leacock, Randee Tengi, and Ross~T. Bunker. 1993.
\newblock \href {https://www.aclweb.org/anthology/H93-1061} {{A Semantic
  Concordance}}.
\newblock In \emph{{H}uman {L}anguage {T}echnology: Proceedings of a Workshop
  Held at Plainsboro}, New Jersey, USA.

\bibitem[{Navigli and Ponzetto(2012)}]{NavigliPonzetto:12aij}
Roberto Navigli and Simone~Paolo Ponzetto. 2012.
\newblock \href
  {https://www.sciencedirect.com/science/article/pii/S0004370212000793}
  {{BabelNet: The Automatic Construction, Evaluation and Application of a
  Wide-Coverage Multilingual Semantic Network}}.
\newblock \emph{Artificial Intelligence}, 193:217--250.

\bibitem[{Neelakantan et~al.(2014)Neelakantan, Shankar, Passos, and
  McCallum}]{neelakantan2014efficient}
Arvind Neelakantan, Jeevan Shankar, Alexandre Passos, and Andrew McCallum.
  2014.
\newblock \href {https://doi.org/10.3115/v1/D14-1113} {{Efficient
  Non-parametric Estimation of Multiple Embeddings per Word in Vector Space}}.
\newblock In \emph{Proceedings of the 2014 Conference on Empirical Methods in
  Natural Language Processing ({EMNLP})}, pages 1059--1069, Doha, Qatar.
  Association for Computational Linguistics.

\bibitem[{Ostrovsky et~al.(2012)Ostrovsky, Rabani, Schulman, and
  Swamy}]{ostrovsky2012effectiveness}
Rafail Ostrovsky, Yuval Rabani, Leonard~J. Schulman, and Chaitanya Swamy. 2012.
\newblock \href {https://dl.acm.org/doi/10.1145/2395116.2395117} {{The
  effectiveness of Lloyd-type methods for the k-means problem}}.
\newblock \emph{Journal of the ACM (JACM)}, 59(6):28.

\bibitem[{Pavlick et~al.(2015)Pavlick, Rastogi, Ganitkevitch, Van~Durme, and
  Callison-Burch}]{pavlick2015ppdb}
Ellie Pavlick, Pushpendre Rastogi, Juri Ganitkevitch, Benjamin Van~Durme, and
  Chris Callison-Burch. 2015.
\newblock \href {https://doi.org/10.3115/v1/P15-2070} {{PPDB 2.0: Better
  paraphrase ranking, fine-grained entailment relations, word embeddings, and
  style classification}}.
\newblock In \emph{Proceedings of the 53rd Annual Meeting of the Association
  for Computational Linguistics and the 7th International Joint Conference on
  Natural Language Processing (Volume 2: Short Papers)}, pages 425--430,
  Beijing, China. Association for Computational Linguistics.

\bibitem[{Peters et~al.(2018)Peters, Neumann, Iyyer, Gardner, Clark, Lee, and
  Zettlemoyer}]{peters2018deep}
Matthew Peters, Mark Neumann, Mohit Iyyer, Matt Gardner, Christopher Clark,
  Kenton Lee, and Luke Zettlemoyer. 2018.
\newblock \href {https://doi.org/10.18653/v1/N18-1202} {{Deep Contextualized
  Word Representations}}.
\newblock In \emph{Proceedings of the 2018 Conference of the North American
  Chapter of the Association for Computational Linguistics: Human Language
  Technologies, Volume 1 (Long Papers)}, pages 2227--2237, New Orleans,
  Louisiana. Association for Computational Linguistics.

\bibitem[{Pilehvar and
  Camacho-Collados(2019)}]{pilehvar-camacho-collados-2019-wic}
Mohammad~Taher Pilehvar and Jose Camacho-Collados. 2019.
\newblock \href {https://doi.org/10.18653/v1/N19-1128} {{{W}i{C}: the
  Word-in-Context Dataset for Evaluating Context-Sensitive Meaning
  Representations}}.
\newblock In \emph{Proceedings of the 2019 Conference of the North {A}merican
  Chapter of the Association for Computational Linguistics: Human Language
  Technologies, Volume 1 (Long and Short Papers)}, pages 1267--1273,
  Minneapolis, Minnesota. Association for Computational Linguistics.

\bibitem[{Pimentel et~al.(2020)Pimentel, Hall~Maudslay, Blasi, and
  Cotterell}]{pimentel-etal-2020-speakers}
Tiago Pimentel, Rowan Hall~Maudslay, Damian Blasi, and Ryan Cotterell. 2020.
\newblock \href {https://doi.org/10.18653/v1/2020.emnlp-main.328} {{Speakers
  Fill Lexical Semantic Gaps with Context}}.
\newblock In \emph{Proceedings of the 2020 Conference on Empirical Methods in
  Natural Language Processing (EMNLP)}, pages 4004--4015, Online. Association
  for Computational Linguistics.

\bibitem[{Radford et~al.(2019)Radford, Wu, Child, Luan, Amodei, and
  Sutskever}]{radford2019language}
Alec Radford, Jeffrey Wu, Rewon Child, David Luan, Dario Amodei, and Ilya
  Sutskever. 2019.
\newblock \href
  {https://cdn.openai.com/better-language-models/language_models_are_unsupervised_multitask_learners.pdf}
  {{Language Models are Unsupervised Multitask Learners}}.
\newblock \emph{OpenAI Blog}, 1(8):9.

\bibitem[{Reif et~al.(2019)Reif, Yuan, Wattenberg, Viegas, Coenen, Pearce, and
  Kim}]{reif2019visualizing}
Emily Reif, Ann Yuan, Martin Wattenberg, Fernanda~B. Viegas, Andy Coenen, Adam
  Pearce, and Been Kim. 2019.
\newblock \href
  {http://papers.nips.cc/paper/9065-visualizing-and-measuring-the-geometry-of-bert.pdf}
  {{Visualizing and Measuring the Geometry of BERT}}.
\newblock In \emph{Advances in Neural Information Processing Systems}, pages
  8592--8600, Vancouver, Canada.

\bibitem[{Reisinger and Mooney(2010)}]{reisinger-mooney-2010-multi}
Joseph Reisinger and Raymond~J. Mooney. 2010.
\newblock \href {https://www.aclweb.org/anthology/N10-1013} {{Multi-Prototype
  Vector-Space Models of Word Meaning}}.
\newblock In \emph{Human Language Technologies: The 2010 Annual Conference of
  the North {A}merican Chapter of the Association for Computational
  Linguistics}, pages 109--117, Los Angeles, California. Association for
  Computational Linguistics.

\bibitem[{Rogers et~al.(2020)Rogers, Kovaleva, and
  Rumshisky}]{rogers2020primer}
Anna Rogers, Olga Kovaleva, and Anna Rumshisky. 2020.
\newblock \href {https://doi.org/10.1162/tacl_a_00349} {{A Primer in
  {BERT}ology: What We Know About How {BERT} Works}}.
\newblock \emph{Transactions of the Association for Computational Linguistics},
  8:842--866.

\bibitem[{Rosenfeld and Erk(2018)}]{rosenfeld-erk-2018-deep}
Alex Rosenfeld and Katrin Erk. 2018.
\newblock \href {https://doi.org/10.18653/v1/N18-1044} {{Deep Neural Models of
  Semantic Shift}}.
\newblock In \emph{Proceedings of the 2018 Conference of the North {A}merican
  Chapter of the Association for Computational Linguistics: Human Language
  Technologies, Volume 1 (Long Papers)}, pages 474--484, New Orleans,
  Louisiana. Association for Computational Linguistics.

\bibitem[{Rousseeuw(1987)}]{rousseeuw1987silhouettes}
Peter~J. Rousseeuw. 1987.
\newblock \href
  {https://www.sciencedirect.com/science/article/pii/0377042787901257?via%3Dihub}
  {{Silhouettes: A graphical aid to the interpretation and validation of
  cluster analysis}}.
\newblock \emph{Journal of computational and applied mathematics}, 20:53--65.

\bibitem[{Schlechtweg et~al.(2020)Schlechtweg, McGillivray, Hengchen,
  Dubossarsky, and Tahmasebi}]{schlechtweg2020semeval}
Dominik Schlechtweg, Barbara McGillivray, Simon Hengchen, Haim Dubossarsky, and
  Nina Tahmasebi. 2020.
\newblock \href {https://www.aclweb.org/anthology/2020.semeval-1.1}
  {{SemEval-2020 Task 1: Unsupervised Lexical Semantic Change Detection}}.
\newblock In \emph{Proceedings of the Fourteenth Workshop on Semantic
  Evaluation}, pages 1--23, Barcelona (online). International Committee for
  Computational Linguistics.

\bibitem[{Su{\'a}rez et~al.(2019)Su{\'a}rez, Sagot, and
  Romary}]{suarez2019asynchronous}
Pedro Javier~Ortiz Su{\'a}rez, Beno{\^\i}t Sagot, and Laurent Romary. 2019.
\newblock \href {https://hal.inria.fr/hal-02148693/document} {Asynchronous
  pipeline for processing huge corpora on medium to low resource
  infrastructures}.
\newblock In \emph{Proceedings of the 7th Workshop on the Challenges in the
  Management of Large Corpora (CMLC-7)}, Cardiff, UK. Leibniz-Institut f{\"u}r
  Deutsche Sprache.

\bibitem[{Talmor et~al.(2020)Talmor, Elazar, Goldberg, and
  Berant}]{talmor2019olmpics}
Alon Talmor, Yanai Elazar, Yoav Goldberg, and Jonathan Berant. 2020.
\newblock \href {https://doi.org/10.1162/tacl_a_00342} {{oLMpics-On What
  Language Model Pre-training Captures}}.
\newblock \emph{Transactions of the Association for Computational Linguistics},
  8:743--758.

\bibitem[{Tenney et~al.(2019)Tenney, Das, and Pavlick}]{tenney-etal-2019-bert}
Ian Tenney, Dipanjan Das, and Ellie Pavlick. 2019.
\newblock \href {https://doi.org/10.18653/v1/P19-1452} {{BERT Rediscovers the
  Classical {NLP} Pipeline}}.
\newblock In \emph{Proceedings of the 57th Annual Meeting of the Association
  for Computational Linguistics}, pages 4593--4601, Florence, Italy.
  Association for Computational Linguistics.

\bibitem[{Vaswani et~al.(2017)Vaswani, Shazeer, Parmar, Uszkoreit, Jones,
  Gomez, Kaiser, and Polosukhin}]{vaswani2017attention}
Ashish Vaswani, Noam Shazeer, Niki Parmar, Jakob Uszkoreit, Llion Jones,
  Aidan~N. Gomez, {\L}ukasz Kaiser, and Illia Polosukhin. 2017.
\newblock \href
  {https://papers.nips.cc/paper/7181-attention-is-all-you-need.pdf} {{Attention
  Is All You Need}}.
\newblock In \emph{Advances in Neural Information Processing Systems}, pages
  5998--6008, Long Beach, California, USA.

\bibitem[{Voita et~al.(2019{\natexlab{a}})Voita, Sennrich, and
  Titov}]{voita-etal-2019-bottom}
Elena Voita, Rico Sennrich, and Ivan Titov. 2019{\natexlab{a}}.
\newblock \href {https://doi.org/10.18653/v1/D19-1448} {{The Bottom-up
  Evolution of Representations in the Transformer: A Study with Machine
  Translation and Language Modeling Objectives}}.
\newblock In \emph{Proceedings of the 2019 Conference on Empirical Methods in
  Natural Language Processing and the 9th International Joint Conference on
  Natural Language Processing (EMNLP-IJCNLP)}, pages 4396--4406, Hong Kong,
  China. Association for Computational Linguistics.

\bibitem[{Voita et~al.(2019{\natexlab{b}})Voita, Talbot, Moiseev, Sennrich, and
  Titov}]{voita-etal-2019-analyzing}
Elena Voita, David Talbot, Fedor Moiseev, Rico Sennrich, and Ivan Titov.
  2019{\natexlab{b}}.
\newblock \href {https://doi.org/10.18653/v1/P19-1580} {{Analyzing Multi-Head
  Self-Attention: Specialized Heads Do the Heavy Lifting, the Rest Can Be
  Pruned}}.
\newblock In \emph{Proceedings of the 57th Annual Meeting of the Association
  for Computational Linguistics}, pages 5797--5808, Florence, Italy.
  Association for Computational Linguistics.

\bibitem[{Vuli{\'c} et~al.(2020)Vuli{\'c}, Ponti, Litschko, Glava{\v{s}}, and
  Korhonen}]{vulic2020probing}
Ivan Vuli{\'c}, Edoardo~Maria Ponti, Robert Litschko, Goran Glava{\v{s}}, and
  Anna Korhonen. 2020.
\newblock \href {https://doi.org/10.18653/v1/2020.emnlp-main.586} {{Probing
  Pretrained Language Models for Lexical Semantics}}.
\newblock In \emph{Proceedings of the 2020 Conference on Empirical Methods in
  Natural Language Processing (EMNLP)}, pages 7222--7240, Online. Association
  for Computational Linguistics.

\bibitem[{Wiedemann et~al.(2019)Wiedemann, Remus, Chawla, and
  Biemann}]{wiedemann2019does}
Gregor Wiedemann, Steffen Remus, Avi Chawla, and Chris Biemann. 2019.
\newblock \href
  {https://www.inf.uni-hamburg.de/en/inst/ab/lt/publications/2019-wiedemannetal-konvens-bert.pdf}
  {{Does {BERT} Make Any Sense? Interpretable Word Sense Disambiguation with
  Contextualized Embeddings}}.
\newblock In \emph{Proceedings of the 15th Conference on Natural Language
  Processing (KONVENS 2019): Long Papers}, pages 161--170, Erlangen, Germany.
  German Society for Computational Linguistics \& Language Technology.

\bibitem[{Wolf et~al.(2020)Wolf, Debut, Sanh, Chaumond, Delangue, Moi, Cistac,
  Rault, Louf, Funtowicz, Davison, Shleifer, von Platen, Ma, Jernite, Plu, Xu,
  Le~Scao, Gugger, Drame, Lhoest, and Rush}]{Wolf2019HuggingFacesTS}
Thomas Wolf, Lysandre Debut, Victor Sanh, Julien Chaumond, Clement Delangue,
  Anthony Moi, Pierric Cistac, Tim Rault, Remi Louf, Morgan Funtowicz, Joe
  Davison, Sam Shleifer, Patrick von Platen, Clara Ma, Yacine Jernite, Julien
  Plu, Canwen Xu, Teven Le~Scao, Sylvain Gugger, Mariama Drame, Quentin Lhoest,
  and Alexander Rush. 2020.
\newblock \href {https://doi.org/10.18653/v1/2020.emnlp-demos.6}
  {{Transformers: State-of-the-Art Natural Language Processing}}.
\newblock In \emph{Proceedings of the 2020 Conference on Empirical Methods in
  Natural Language Processing: System Demonstrations}, pages 38--45, Online.
  Association for Computational Linguistics.

\bibitem[{Zhang(2001)}]{zhang2001dependence}
Bin Zhang. 2001.
\newblock \href {https://www.hpl.hp.com/techreports/2001/HPL-2001-91.html}
  {{Dependence of Clustering Algorithm Performance on Clustered-ness of Data}}.
\newblock \emph{HP Labs Technical Report HPL-2001-91}.

\bibitem[{Zhou et~al.(2019)Zhou, Ge, Xu, Wei, and Zhou}]{zhou-etal-2019-bert}
Wangchunshu Zhou, Tao Ge, Ke~Xu, Furu Wei, and Ming Zhou. 2019.
\newblock \href {https://doi.org/10.18653/v1/P19-1328} {{{BERT-based Lexical
  Substitution}}}.
\newblock In \emph{Proceedings of the 57th Annual Meeting of the Association
  for Computational Linguistics}, pages 3368--3373, Florence, Italy.
  Association for Computational Linguistics.

\bibitem[{Zhu et~al.(2015)Zhu, Kiros, Zemel, Salakhutdinov, Urtasun, Torralba,
  and Fidler}]{Zhuetal2015}
Yukun Zhu, Ryan Kiros, Rich Zemel, Ruslan Salakhutdinov, Raquel Urtasun,
  Antonio Torralba, and Sanja Fidler. 2015.
\newblock \href {https://doi.org/10.1109/ICCV.2015.11} {{Aligning Books and
  Movies: Towards Story-Like Visual Explanations by Watching Movies and Reading
  Books}}.
\newblock In \emph{Proceedings of the 2015 IEEE International Conference on
  Computer Vision (ICCV'15)}, page 19–27, Santiago, Chile. IEEE Computer
  Society.

\bibitem[{Zipf(1945)}]{Zipf1945}
George~Kingsley Zipf. 1945.
\newblock The meaning-frequency relationship of words.
\newblock \emph{Journal of General Psychology}, 33(2):251–256.

\end{thebibliography}
\bibliographystyle{acl_natbib}

\appendix

\section{Clusterability Metrics} \label{app:metrics}

\noindent \textbf{Variance Ratio}. First, the variance of a cluster $y$ is calculated:
 
 \vspace{-2mm}    
\begin{equation}
\sigma^2(Y) = \frac{1}{|y|} \sum_{i \epsilon y} (y_{i} - \bar{y})^2
\end{equation}

\noindent where $\bar{y}$ denotes the centroid of cluster $y$. Then the within-cluster variance $W$ and the between-cluster variance $B$ of a clustering solution $C$ are calculated in the following way:

\vspace{-3mm}   

\begin{equation}
 W(C) = \sum_{j=1}^{k}{p_{j}} \sigma^2(x_{j}) \\ 
\end{equation}

\begin{equation}
  B(C) = \sum_{j=1}^{k} p_{j} (\bar{x}_{j} - \bar{x})^2
\end{equation}

\vspace{-2mm}

\noindent where $x$ is the set of all data points and $p_{j} =  \frac{|x_{j}|}{|x|}$. $x_{j}$ are the data points in cluster $j$.
\noindent Finally, the {\sc VR} of a clustering $C$ is obtained as the ratio between $B(C)$ and $W(C)$:

\vspace{-3mm}    

\begin{equation}
    VR = \frac{B(C)}{W(C)}
\end{equation}

\noindent \textbf{Separability} ({\sc sep}). In an optimal clustering $C_{k}$ of the dataset $x$ with $k$ clusters, {\sc sep} is defined as follows:

\vspace{-4mm}    
\begin{equation}
SEP(x,k) = \frac{loss(C_{k})}{loss(C_{k-1})} 
\end{equation}

\end{document}